%% file: 088-main.tex
\documentclass[runningheads]{llncs}
\RequirePackage{silence}  
\usepackage{graphicx}
\usepackage{comment}
\usepackage{amsmath,amssymb}
\usepackage{color}
\usepackage{url}
\usepackage{hyperref}

\usepackage{graphicx}
\usepackage{booktabs}
\usepackage{subcaption}
\usepackage[table]{xcolor}

\usepackage{float}
\usepackage{orcidlink}
\newcommand{\X}{\mathbf{X}}
\newif\ifreview
\reviewtrue
\reviewfalse

\ifreview
	\usepackage{lineno}

	\linenumbers
\fi

\begin{document}

%%%%%%%%%%%%%%%%%%%%% Add submission id, track, and title. %%%%%%%%%%%%%%%%%%%%%

% TODO: Please insert your submission number here
\def\SubNumber{88}

% TODO: Please uncomment the track this paper will be submitted to, comment all other lines
%\def\GCPRTrack{Main Track}
%\def\GCPRTrack{Special Track: Pattern recognition in the life and natural sciences}
%\def\GCPRTrack{Special Track: Photogrammetry and remote sensing}
%\def\GCPRTrack{Special Track: Computer vision systems and applications}
%\def\GCPRTrack{Young Researcher's Forum}
\def\GCPRTrack{Fast Review Track}
%\def\GCPRTrack{Extended Abstract}

% TODO: Replace with your title
\title{Multi-Person Human Motion Forecasting in Complex Scenes}
% You can use \thanks for acknowledgment. Do not add any acknowledgment to the draft 
% version that is used for the review process.  
%\title{Title\thanks{XXX}}

\ifreview
	% ANONYMOUS SUBMISSION FOR REVIEW
	% DO NOT MODIFY these for the draft version that is used for the review process.
	\titlerunning{GCPR 2026 Submission \SubNumber{}. CONFIDENTIAL REVIEW COPY.}
	\authorrunning{GCPR 2026 Submission \SubNumber{}. CONFIDENTIAL REVIEW COPY.}
	\author{GCPR 2026 - \GCPRTrack{}}
	\institute{Paper ID \SubNumber}
\else

%	\author{Serdar  Ozsoy \inst{1}\orcidID{0000-0003-1765-4708} \and
%	Lars Doorenbos\inst{1,2}\orcidID{0000-0002-0231-9950} \and
%	Juergen Gall\inst{1,2}\orcidID{0000-0002-9447-3399}}
    
	\author{Serdar  Ozsoy \inst{1,2}\thanks{Corresponding Author}\orcidlink{0000-0003-1765-4708}  \and
	Lars Doorenbos\inst{1,2}\orcidlink{0000-0002-0231-9950} \and
	Juergen Gall\inst{1,2}\orcidlink{0000-0002-9447-3399}}

	\authorrunning{S. Ozsoy et al.}
	
	\institute{University of Bonn, Germany \and Lamarr Institute for Machine Learning and Artificial Intelligence, Germany
	%\url{http://www.springer.com/gp/computer-science/lncs} \and ABC Institute, Rupert-Karls-University Heidelberg, Heidelberg, Germany\\
	\email{\{soezsoy,doorenbos,gall\}@iai.uni-bonn.de}\\
    %\url{https://github.com/serdarozsoy/OCSD}
    \url{https://serdarozsoy.github.io/OCSD-project/}
    }

\fi

\maketitle              % typeset the header of the contribution

\begin{abstract}
Accurately forecasting the movement of people in complex scenes requires reasoning over the past and present state of the entire environment. In this context, effectively incorporating object information and social interactions into a unified framework remains particularly challenging. To address this, we propose Object-Conditioned Social Diffusion (OCSD), a conditional diffusion model that integrates motion history, multi-person interactions, and object cues into a single framework. OCSD uses an object-conditioning mechanism that modulates denoising at every timestep, enabling fine-grained human-object reasoning, and a social encoder that models the interactions between all humans in the scene. As a result, our model naturally handles varying group sizes, complex social interactions, and supports sampling multiple plausible futures. Extensive experiments show that OCSD achieves state-of-the-art results on the Humans in Kitchens (HiK) and HOI-M\textsuperscript{3} benchmarks. It reduces the two-second path error by 121.5 mm (31.3\%) on HiK and 130.5 mm (33.2\%) on HOI-M\textsuperscript{3} compared to prior work, and produces more realistic long-term forecasts.
%We will release our code upon acceptance. 
%Our code is available at: \url{https://github.com/serdarozsoy/OCSD}
  \keywords{Human motion forecasting \and Diffusion models}
\end{abstract}

\section{Introduction}
\label{sec:intro}

Anticipating how people will move in complex scenes is a crucial capability for autonomous systems to operate safely in various domains, including robotics \cite{foka2010probabilistic,fridovich2020confidence,unhelkar2018human} and autonomous driving \cite{8550300,li2021attentional,li2020socially,scholler2020constant}.
However, achieving the accuracy desired to do so is a challenging task. 
A key part of this challenge lies in understanding the factors that influence human motion at different timescales. Even for short time horizons, accurate human motion forecasting requires more than extrapolating past poses: models must reason jointly about their own history, interactions with other people in the scene, as well as the objects. Longer time horizons introduce even more complexity for forecasting. Specifically, the mapping from observed motion to future behavior of people in complex scenes becomes an inherently ambiguous problem, where multiple motions can be correct given the same context. 

Short- and long-horizon forecasting are typically considered separately. For instance, short-term works tend to disregard the probabilistic characteristic of human motion and output only the most likely prediction (e.g.,~\cite{sun2025humof}). In contrast, works focusing on longer horizons produce stochastic predictions with generative models, but, as a result, tend to underperform deterministic models in terms of local realism~\cite{mueller2024sast}. Capturing these different factors within a unified framework remains an open problem.
\begin{figure}[t]
    \centering
    \includegraphics[width=0.95\linewidth]{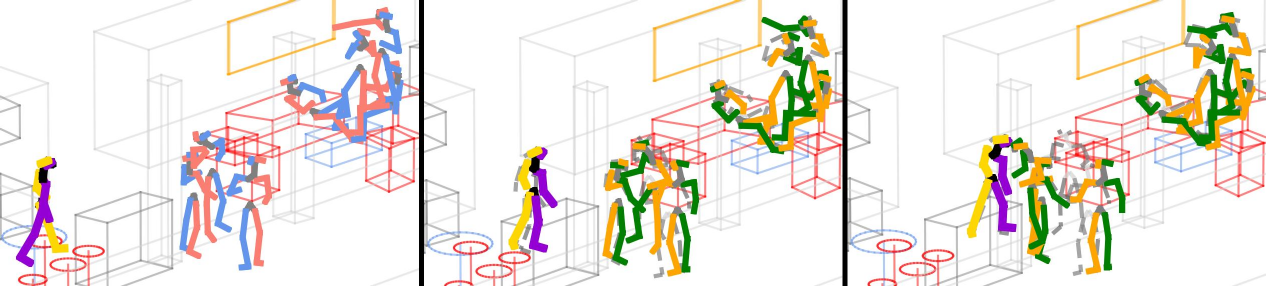}
    \caption{\textbf{Multi-person human motion forecasts in complex scenes with OCSD.} We show the last observed frame on the left. The next two panels display forecast poses after 1 second (middle) and 2 seconds (right), with grey skeletons denoting the ground truth and colored skeletons representing the predictions. The gray skeletons are well aligned with the forecast poses, which shows that our model can generate plausible poses in complex scenes by integrating context.
    }
    \label{fig:teaser}
\end{figure}

To address this challenge, we propose a conditional denoising diffusion model that holistically integrates the entire scene context.
Our model uses an object-conditioning mechanism that modulates the denoising at each timestep, through which objects can impact the predicted future motion. Additionally, we use a social encoder that extracts the social context and integrates it into our model. 
As a result, our model can capture fine-grained human-object dependencies while also regarding social context, as shown in Fig.~\ref{fig:teaser}. 

We show through extensive experiments that these modeling capabilities translate into strong empirical performance. We evaluate our method on two benchmarks, Humans in Kitchens (HiK)~\cite{tanke2023humans} and HOI-M\textsuperscript{3}~\cite{zhang2024hoi}, on both short- and long-term human motion forecasting. For short horizons, the sequences our model generates outperform those from eight baselines in accuracy of predicted paths and poses, while for longer horizons our method reaches high realism and diversity. For example, on HiK, which has the most people in the scene, we reduce the state-of-the-art two-second path error by $121.5$mm, representing an improvement of $31.3$\%. For longer forecasts up to ten seconds, our model achieves $25.9$\% higher realism than the previous state-of-the-art approaches. Similarly, our model reduces the two-second path error by 130.5 mm (33.2\%) on HOI-M\textsuperscript{3} compared to prior work.
In short, our main contributions are as follows:
\begin{itemize}
\item We introduce OCSD, a conditional diffusion framework that unifies past conditioning, social context, and object signals for multi-person motion forecasting in complex scenes.
\item We set a new state-of-the-art across various metrics on the HiK and HOI-M\textsuperscript{3} datasets.
\end{itemize}

\section{Related Works}
\label{sec:related}

\hspace{\parindent}\textbf{Single-Person Human Motion Prediction.}
Early works in human motion prediction operate in single-person settings, with both deterministic models that predict a single future sequence and probabilistic models that predict multiple diverse future motions. Deterministic methods typically use Recurrent Neural Networks~\cite{fragkiadaki2015recurrent,jain2016structural,martinez2017human}, Graph Convolutional Networks~\cite{ma2022progressively,mao2020history,mao2019learning}, Transformers~\cite{aksan2021spatio,cai2020learning}, or Multilayer Perceptron-based~\cite{bouazizi2022motionmixer,guo2023back,wei2019motion} approaches, while Generative Adversarial Networks~\cite{barsoum2018hp,kundu2019bihmp} and
Variational Autoencoders~\cite{yan2018mt} form the foundations for probabilistic human motion prediction.
Subsequent work in probabilistic prediction mostly focuses on increasing the diversity of the predictions~\cite{aliakbarian2021contextually,aliakbarian2020stochastic,dang2022diverse}. In addition, \cite{mao2021generating,xu2024learning,xu2022diverse,yuan2020dlow} not only diversify predictions but also offer a way to control the diversity of predicted motions. Recently, diffusion-based models \cite{barquero2023belfusion,chen2023humanmac,curreli2025nonisotropic,saadatnejad2023diffusion,sun2024comusion} achieve high-fidelity forecasts while providing diversity, for instance, by re-framing the prediction as a masked motion completion task~\cite{chen2023humanmac}, or introducing latent diffusion~\cite{barquero2023belfusion}.

\textbf{Multi-Person Human Motion Prediction.}
The setting of multi-person human motion prediction introduces an extra  challenge of modeling interdependent social dynamics. Transformer-based architectures \cite{jeong2024multi,peng2023trajectory,vendrow2022somoformer,wang2021multi,xiao2024multi,xu2023joint} became a common approach  in this domain. For example, \cite{wang2021multi} uses a local-range encoder for individual motion and a global-range encoder for social interactions. Subsequent work refined this by modeling interactions at a finer granularity. For instance, \cite{peng2023trajectory} proposes modeling interactions at the body-part level, while \cite{xu2023joint} introduces an auxiliary loss to predict the future distance between joints. 
Besides transformer-based models, \cite{zheng2025efficient} proposes computationally efficient modeling of social interaction via an MLP-based framework. Generative models are also used in this context to capture social stochasticity: \cite{xu2022stochastic} introduces a dual-level generative modeling framework to separately model independent individual motion and global social interactions, and~\cite{Tanke_2023_ICCV} leverages a diffusion model with an order-invariant aggregation function to flexibly manage a varying number of people.

\textbf{Human Motion Prediction with Object Context.}
Integrating object information is typically considered separate from social context and has its own set of approaches.
As an early method, \cite{mao2022contact} proposes a two-stage pipeline to first predict future contact maps from a scene point cloud, and then forecasts the pose conditioned on these maps. \cite{scofano2023staged} builds on this with a three-stage pipeline by predicting contact points, root trajectory, and full pose, respectively. While \cite{xing2024scene} proposes to constrain the whole-body motion by introducing the Signed Distance Function (SDF) volume as a dense global scene representation, \cite{wang2024harmonizing} uses segmented object instances from 3D point clouds to predict diverse human motion.

Only recently, unified models attempt to merge both social and scene context. \cite{adeli2021tripod} is an early attempt using GNNs with a pretrained object detector. \cite{mueller2024sast} introduces a diffusion-based model for multi-person settings that uses scene object information while targeting long-term (up to 10s) generation with an emphasis on diversity and plausibility. In contrast, \cite{sun2025humof} also offered a unified hierarchical framework that operates on 
scene objects while focusing on more short-term (0.5 - 2s) prediction accuracy. Our model bridges these two directions: we achieve state-of-the-art short-term (0.5-2s) path and pose accuracy while also ensuring long-term (10s) plausibility and diversity with both object and social context.

\section{Method}
\label{sec:method}

Forecasting the motion of multiple people in complex scenes requires modeling several factors: the past motion, the interactions between people in the scene, and the type and location of various objects.
We denote the observed motion sequence of a group of $P$ people for a duration of $T_{in}$ frames by a tensor of coordinates $\mathbf{X}_{T_{in}} \in \mathbb{R}^{T_{in} \times P \times J \times 3}$, where $J$ represents the number of joints, and we refer to the pose of a single person $p$ at frame $t$ as $\mathbf{x}_t^{(p)}$. If there are fewer than $P$ people in the scene, the superfluous dimensions are filled with zeros. The object information is encoded as a set of $M$ objects $\mathbf{O} = \{o_1,\cdots,o_M\}$, with each object $o$ represented by $N$ vertices $\mathbf{V}_o \in \mathbb{R}^{N\times3}$, fixed at the last frame of the observations, and a semantic type $y$. 
Given this context, the goal is to forecast the poses of all people in the future $T_{out}$ frames, $\mathbf{X}_{T_{out}} \in \mathbb{R}^{T_{out} \times P \times J \times 3}$, conditioned on the observed sequences $\X_{T_{in}}$ and the object information $\mathbf{O}$. The full sequence thus lasts $T=T_{in} + T_{out}$ frames.

\begin{figure}[!ht]
    \centering
    \includegraphics[width=0.95\linewidth]{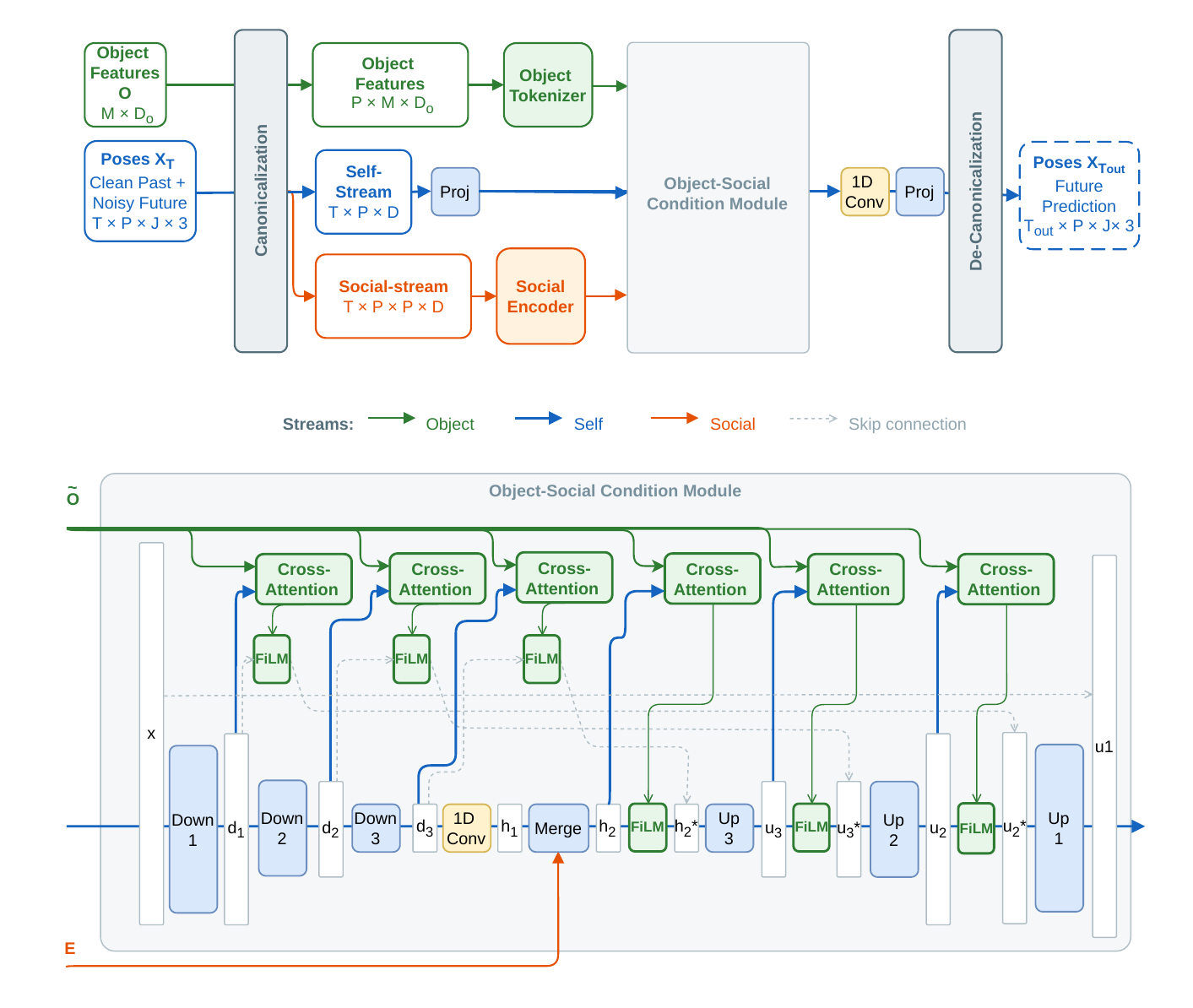}
    \caption{\textbf{Overview of OCSD.} \textit{Top:} Our method takes a sequence of poses for a dynamic number of people in the scene, along with object information. These inputs are split into the \textit{self-stream} and two conditioning streams for the objects and social context. From this, OCSD predicts the poses of all people for the next $T_{out}$ frames. \textit{Bottom:} Details of our main model, following a U-Net architecture. We integrate the social context for each reference person into the bottleneck, while object information is added throughout the network via FiLM modules. 
}
    \label{fig:arch}
\end{figure}

Learning this mapping from observed motion to future behavior of people in complex scenes is an inherently ambiguous problem: for the same input, multiple plausible motions can be correct, especially for longer-term predictions. Therefore, we frame the motion forecasting task as learning a conditional distribution of future motion based on the observed motion and scene context, for which we rely on conditional diffusion models~\cite{ho2020denoising}.
The core of our model, OCSD, lies in how it represents and fuses information, which we describe in the sections below. 
We provide an overview of our method in Fig.~\ref{fig:arch}.

\subsection{Input Representations}
\label{sec:repr}

\hspace{\parindent}\textbf{Pose representation.}
The observed poses $\X_{T_{in}}$ contain the joint coordinates in the global coordinate system. However, directly operating in the global coordinates is suboptimal for pose forecasting~\cite{Guo_2022_CVPR}. Therefore, we use an alternative representation as in~\cite{Tanke_2023_ICCV}, and transform each per-person pose vector by
$\mathbf{x}_t^{(p)} = \left[ \mu_t^{(p)} \in \mathbb{R}^3, r_t^{(p)} \in \mathbb{R}^3, q_t^{(p)} \in \mathbb{R}^{3J} \right]$.
Here, $\mu_t^{(p)}$ is the global translation given by the 3D world-coordinate position of the hip center, and $r_t^{(p)}$ is the global rotation of the body's 2D-plane facing direction represented as a 3D rotation. $q_t^{(p)}$ are the local joint positions, representing the 3D positions of the $J$ joints where the person is translated to the origin and rotated such that the hips are parallel to the x-axis. 
The total dimensionality of $\mathbf{x}_t^{(p)}$, therefore, is $D=6+3J$. We further standardize the local joints by the mean and standard deviation of each joint in the training set.

\textbf{Social context representation.}
After this processing, all pose vectors are represented in their own local coordinate system. While this is important for stable learning \cite{sun2017human}, it destroys all information about the position of a person relative to the other people in the scene. 
To use this information in the model, we design a social representation that considers the relative position of other people in the scene from the perspective of each person.
In this context, we refer to the person whose perspective is considered as the \textit{primary person} $p^\star$. 

To obtain the social information for $p^\star$, we establish a translation $\mathbf{T}^{(p^\star)}$ and rotation $\mathbf{R}^{(p^\star)}$ with regard to their pose in the last observed frame $t_{ref}$. Then, for every %other 
person $p$, we transform their entire pose vector $\mathbf{x}_t^{(p)}$ into this coordinate system.
Using these pose vectors, we construct a pairwise tensor $\boldsymbol{\mathcal{X}}\in\mathbb{R}^{T\times P\times P\times D}$ where each slice $\boldsymbol{\mathcal{X}}_{t}^{(p^\star,p)}$ contains %$\widetilde{\mathbf{x}}_t^{(p)}$, i.e., 
$\mathbf{x}_t^{(p)}$ from the perspective of $p^\star$. 
From $\boldsymbol{\mathcal{X}}$, we extract two separate representations.
These are the \textit{self-stream}, representing the pose vector of all individual people in the scene (i.e., the diagonal of $\boldsymbol{\mathcal{X}}$), and the \textit{social-stream}, which contains pose vectors of all other people in the scene relative to the reference person $p^{\star}$.
The \textit{self-stream} is directly fed into the U-Net~\cite{ronneberger2015u} of $f_\theta$. In contrast, the \textit{social-stream} will be used to control the generation process, as described in Sec.~\ref{sec:context}.

\textbf{Object context representation.} 
To represent the objects present in the scene, we create personalized object information representations in a similar way to the social representation, encoding the relationship between a reference person and the objects in the scene.
We canonicalize the vertices of an object $\mathbf{V}_o$ into each person's $p^\star$ reference frame using the same translation $\mathbf{T}^{(p^\star)}$ and rotation $\mathbf{R}^{(p^\star)}$ 
as in social context.
Besides flattening N vertices to 3N points, we encode the semantics of the object into a one-hot vector. These are then combined into the object representations $\mathbf{\tilde{V}}$ by concatenation, leading to a final representation of $D_o = 3N + C$, with $C$ the number of object types present.

\subsection{Integrating Social and Object Context}
\label{sec:context}
 
We propose to use two mechanisms to integrate the social and object context representations into the pose forecasting model $f_\theta$ to control the generation process, which we show schematically in Fig.~\ref{fig:arch} and describe in turn below.

\textbf{Social integration.}
The social context is represented in the \textit{social-stream} obtained from the pairwise relational tensor $\boldsymbol{\mathcal{X}}$. Similar to~\cite{Tanke_2023_ICCV}, we pass this information through a learned encoder and integrate the resulting embedding in the bottleneck of the U-Net.
We utilize this Social Encoder to temporally downsample the \textit{social-stream} through three successive blocks, each consisting of three strided 1D convolutions with a SiLU activation function and layer normalization. This encoder outputs a set of features $\mathbf{E} \in \mathbb{R}^{T/8\times P\times P \times d'}$
, and $\mathbf{e}^{(p^\star, p)}$ denotes the encoded representation for person $p$ relative to primary person $p^\star$.

One challenge to consider is that there can be a dynamic number of people present in the scene. Therefore, to obtain a single and fixed-dimensional social representation for a primary person, $\mathbf{e}^{(p^\star)}$, we mask out any zero-padded entries of $E$ that are not related to any person. 
Then, we perform mean pooling across the second $P$ dimension. 
The final vector $\mathbf{e}^{(p^\star)}$ summarizes the entire social context from the perspective of $p^\star$ and is concatenated to the $p^\star$'s \textit{self-stream} features at the U-Net bottleneck, followed by a 1D Convolution, denoted as ``Merge" in Fig. \ref{fig:arch}. Thereby, the model jointly processes the primary person's own motion while integrating the surrounding social dynamics.

\textbf{Object context integration.} We design our object conditioning as a multi-scale modulation mechanism applied at multiple stages $s$, which correspond to the features H $\in$ [$d_{1}$, $d_{2}$, $d_{3}$, $h_{2}$, $u_{3}$, $u_{2}$] of the U-Net. This process is done for each person $p$ and consists of three parts: First, the canonicalized object vertices are projected via an MLP to match the feature dimension of that stage. The resulting features form a set of $M$ object tokens for each person $p$, which encode local scene information and can be integrated into the model. 
Second, the motion features $H^{(p, s)}_t$ for person $p$ at layer $s$ for each timeframe $t \in T$ are processed by cross-attention modules. 
In this setting, $H^{(p, s)}_t$ acts as the query, and the $M$ projected object tokens serve as keys and values. This way, the attention mechanism produces an embedding that summarizes the objects most relevant to the motion of the person $p$ in frame $t$ at stage $s$.
Finally, we integrate these features via Feature-wise Linear Modulation (FiLM) \cite{perez2018film}. The object embedding
is passed through a linear projection to predict a per-timeframe scale vector $\gamma^{(p,s)}_t$ and a bias vector $\beta^{(p,s)}_t$ to modulate the motion features:
\begin{equation}
    \widehat{H}^{({p,s})}_t \;=\; H^{({p,s})}_t \odot \big(1+\gamma^{({p,s})}_t\big) + \beta^{({p,s})}_t.
\end{equation}
In the first half of the U-Net, FiLM modulation is applied through the skip connections. In contrast, we apply it directly to the features in the second half. This process allows object context to precisely influence the motion generation.

\subsection{Conditional diffusion model}
\label{sec:ddpm}
\hspace{\parindent}\textbf{Training.}
Our diffusion model $f_\theta$ is trained to predict the clean motion $\mathbf{X}_0$ from a noised sample $\mathbf{X}_\tau$ based on the past motion and scene context.
Let $\mathbf{X}_0$ be the full clean motion sequence, which includes the past $\mathbf{X}_{T_{in}}$ and future $\mathbf{X}_{T_{out}}$. Given a diffusion noise schedule $\{\alpha_\tau\}_{\tau=1}^{K}$ with $K$ denoising steps and its cumulative product $\bar\alpha_\tau=\prod_{s=1}^{\tau}\alpha_s$, we form the input tensor $\mathbf{X}_\tau$ by combining the clean past and the noised future: 
\begin{equation}
\X_{\tau}
=
\big[\, \X_{T_{in}} \;;\; \sqrt{\bar\alpha_\tau}\,\X_{T_{out}} + \sqrt{1-\bar\alpha_\tau}\,\varepsilon \,\big],
\end{equation}
where $\varepsilon\sim\mathcal{N}(0,I)$. Our network $f_\theta$ is trained to predict the full clean motion $\mathbf{X}_0$ from $\mathbf{X}_\tau$, conditioned on the denoising step $\tau$ and object context $\mathbf{O}$ with a $\mathcal{L}_{1}$ reconstruction objective:
\begin{equation}
\mathcal{L}_{x_0} 
=
\mathbb{E}_{\tau}\!\left[
\big\|\, \X_0 - f_\theta\!\big(\X_\tau,\, \tau,\, \mathbf{O} \big) \big\|_{1}
\right].
\end{equation}
We modify this base loss by applying a frame mask $\mathbf{M}_{\text{fra}}$ to compute the loss only for frames $t > T_{\text{in}}$
, and a presence mask $\mathbf{M}_{\text{pres}}$ to ignore absent individuals. Moreover, we decompose the masked loss into three components in our hybrid pose representation to ensure stable training: root translation $\mu$, root orientation $r$, and local joint pose $q$. Each is normalized by its number of valid elements:
\begin{equation}
\label{eq:loss}
\mathcal{L}
=
\frac{\|\Delta_{\mathrm{\mu}}\|_{1}}{N_{\mathrm{\mu}}}
+
\frac{\|\Delta_{\mathrm{r}}\|_{1}}{N_{\mathrm{r}}}
+
\frac{\|\Delta_{\mathrm{q}}\|_{1}}{N_{\mathrm{q}}}.
\end{equation}
Here, $\Delta_{\mu}$, $\Delta_{r}$, and $\Delta_{q}$ denote the masked per-element errors for root translation, root orientation, and local joint pose, respectively, and $N_{\mu}$, $N_{r}$, and $N_{q}$ are the valid element count for each component.

\textbf{Inference.}
We sample using the standard DDPM reverse process. To ensure the generated future is coherent with the given past $\mathbf{X}_{T_{in}}$, we apply an inpainting strategy. A single noise vector $\mathbf{z}_{\mathrm{past}}$ is drawn for the past frames, and at each denoising step $\tau$, a noised version of the past is concatenated with the predicted $\mathbf{X}_{\tau-1}$ before passing it to the next step:
\begin{equation}
\X_{\tau-1} \;=\; [\sqrt{\bar\alpha_{\tau}}\,\mathbf{X}_{T_{in}}  + \sqrt{1-\bar\alpha_{\tau}}\,\mathbf{z}_{\mathrm{past}} \,;\, \mathbf{X}_{\tau-1}^{\text{predicted}}]
\end{equation}
which prevents the observed past from drifting during the sampling process and avoids discontinuities between observed and predicted poses.

\section{Experiments}
%We describe datasets, metrics and implementation details briefly below and provide more details in Sec.~\ref{sec:extdatasets}, \ref{sec:extmetrics} and \ref{sec:extimplement} of the supplementary material, respectively.
We describe datasets, metrics and implementation details briefly below and provide more details in the supplementary material.

\label{sec:exp}
\textbf{Datasets.}
We evaluate our method on two recent multi-person human motion forecasting datasets.
\textbf{Humans in Kitchens}~\cite{tanke2023humans} (HiK) is a motion-capture dataset of people interacting in four kitchen environments, containing up to 16 people and 50 annotated objects in a scene. Following the standard protocol \cite{Tanke_2023_ICCV,mueller2024sast,sun2025humof}, we train on three kitchens (A, B, and C) and evaluate on curated sequences from kitchen D, focusing on transitional moments. 
\textbf{HOI-M\textsuperscript{3}}~\cite{zhang2024hoi} is a large-scale multi-human multi-object dataset captured in realistic indoor rooms. It contains 199 long sequences with up to 5 people and at least 5 objects per scene. We use the 49 living room sequences publicly available (1-51, except non-released 43 and 45), and set aside $20\%$ as the test set following~\cite{sun2025humof}.

\textbf{Metrics.} We assess our model's performance using four metrics.
We evaluate short-term forecasts as in~\cite{sun2025humof} with \textbf{Path error}, i.e., the error of root joint trajectory in millimeters (mm), and  \textbf{Pose error}, which is the Mean Per Joint Position Error.
Long-term predictions are scored with \textbf{Normalized Directional Motion Similarity}~\cite{tanke2021intention} (NDMS), which measures the realism of generated motion, and \textbf{Unique Motion Word Ratio}~\cite{mueller2024sast} (UMWR), which tests diversity.

\textbf{Implementation Details.}
We train OCSD for $90$ epochs using the Adam~\cite{kingma2014adam} optimizer with a learning rate of $3 \times 10^{-4}$. We define one epoch as 20,000 training samples. 
We follow the protocol from \cite{mueller2024sast,sun2025humof} and use the HiK dataset at its original 25 FPS. The HOI-M\textsuperscript{3} dataset is downsampled from 60 FPS to 30 FPS. For short-term experiments, we provide 1 second of motion as input to predict the next 2 seconds. For long-term experiments, we use 2 seconds of input to predict 10 seconds.
For evaluation, as our method is stochastic, we forecast four pose sequences per input motion, compute the corresponding metrics, and report the average performance.

\begin{table*}[t]
\centering
\caption{\textbf{Short-term pose forecasting results.} Baseline results are taken from~\cite{sun2025humof}. Our approach consistently reaches the best results.}
\label{tab:main}
\resizebox{0.9\textwidth}{!}{
\begin{tabular}{ll rrrrr r rrrrr}
\hline
Dataset & Method & \multicolumn{5}{c}{Path Error (mm)} & & \multicolumn{5}{c}{Pose Error (mm)} \\
\cline{3-7} \cline{9-13}
& & 0.5s & 1.0s & 1.5s & 2.0s & mean & & 0.5s & 1.0s & 1.5s & 2.0s & mean \\
\hline
{HIK~\cite{tanke2023humans}} 
& ContAware~\cite{mao2022contact} & 138.3 & 251.3 & 352.4 & 430.8 & 239.3 & & 87.8 & 117.1 & 136.1 & 147.8 & 106.8 \\
& GIMO~\cite{zheng2022gimo} & 143.0 & 259.7 & 384.2 & 487.3 & 258.6 & & 85.5 & 121.6 & 142.0 & 153.0 & 109.3 \\
& STAG~\cite{ma2022progressively} & 124.7 & 245.4 & 352.4 & 479.2 & 239.7 & & 81.7 & 110.9 & 132.5 & 140.9 & 100.6 \\
& MutualDistance~\cite{xing2024scene} & 128.7 & 253.2 & 372.5 & 479.2 & 246.0 & & 82.9 & 117.2 & 138.5 & 148.2 & 105.9 \\
& T2P~\cite{jeong2024multi} & 88.6 & 199.6 & 318.8 & 447.1 & 208.7 & & 74.2 & 108.6 & 127.5 & 142.6 & 96.9 \\
& IAFormer ~\cite{xiao2024multi} & 83.9 & 195.0 & 311.1 & 434.9 & 200.1 & & 71.5 & 106.5 & 125.9 & 137.7 & 95.0 \\
& SAST~\cite{mueller2024sast}  & 86.7 & 187.4 & 284.9 & 398.1 & 189.0 & & 72.3 & 101.4 & 118.0 & 128.6 & 93.2 \\
& HUMOF~\cite{sun2025humof}& \textbf{78.8} & 177.4 & 278.8 & 388.4 & 180.7 & & 71.2 & 100.6 & 116.9 & 127.1 & 90.2 \\
\rowcolor{blue!20!white}& \textbf{OCSD  }& 78.9 & \textbf{132.7} & \textbf{197.8} & \textbf{266.9} & \textbf{137.6} & & \textbf{69.8} & \textbf{99.4} & \textbf{112.3} & \textbf{121.9} & \textbf{88.4} \\
\hline
{HOI-M\textsuperscript{3}~\cite{zhang2024hoi}} 
& ContAware~\cite{mao2022contact} & 125.6 & 239.9 & 285.4 & 432.9 & 236.9 & & 106.2 & 152.8 & 174.3 & 197.1 & 137.5 \\
& GIMO~\cite{zheng2022gimo} & 131.4 & 247.7 & 300.9 & 454.4 & 255.2 & & 107.9 & 155.9 & 182.6 & 207.1 & 141.0 \\
& STAG~\cite{ma2022progressively} & 128.1 & 234.4 & 289.5 & 438.1 & 239.7 & & 102.5 & 145.0 & 167.1 & 185.6 & 131.2 \\
& MutualDistance~\cite{xing2024scene}  & 83.6 & 169.7 & 278.8 & 402.8 & 189.9 & & 94.4 & 137.1 & 158.2 & 181.3 & 125.3 \\
& T2P~\cite{jeong2024multi} & 74.2 & 168.8 & 296.9 & 429.2 & 194.1 & & 88.0 & 135.8 & 160.9 & 183.2 & 124.6 \\
& IAFormer~\cite{xiao2024multi}  & 69.0 & 166.6 & 290.1 & 423.5 & 186.3 & & 86.1 & 135.0 & 165.9 & 180.7 & 121.6 \\
& SAST~\cite{mueller2024sast} & 75.0 & 166.2 & 280.4 & 403.9 & 184.8 & & 89.2 & 133.8 & 167.0 & 182.9 & 122.3 \\
& HUMOF~\cite{sun2025humof} & 67.1 & 156.6 & 268.4 & 393.1 & 174.6 & & 86.3 & 129.6 & 155.0 & 172.1 & 117.9 \\
\rowcolor{blue!20!white}& \textbf{OCSD } & \textbf{52.4} & \textbf{115.5} &  \textbf{185.9} &  \textbf{262.6} & \textbf{123.5} &  &\textbf{63.8} & \textbf{93.1} & \textbf{112.5}  & \textbf{127.9} & \textbf{87.3} \\
\hline
\end{tabular}
}
\end{table*}

\begin{table}[t]
\centering
\caption{\textbf{Long-term pose forecasting results on HiK.} Baseline results are taken from~\cite{mueller2024sast}. Unlike previous approaches, our method performs well even for a 10-second horizon.}
\label{tab:comparisons1}
\resizebox{0.5\textwidth}{!}{
\begin{tabular}{l c ccccc}
\toprule
& {NDMS $\uparrow$} & \multicolumn{5}{c}{UMWR $\uparrow$} \\
\cmidrule(l){3-7} % Creates a partial rule under the UMWR columns
& & 2s & 4s & 6s & 8s & 10s \\
\midrule
MRT~\cite{wang2021multi} & 0.16 & 0.09 & 0.08 & 0.07 & 0.07 & - \\
HisRep~\cite{mao2020history} & 0.23 & 0.12 & 0.07 & 0.06 & 0.06 & 0.06 \\
SiMLPe~\cite{guo2023back} &0.27 & 0.15 & 0.07 & 0.07 & 0.06 & 0.06 \\
TriPod~\cite{adeli2021tripod} & 0.13 & 0.18 & 0.14 & 0.12 & 0.11 & 0.11 \\
SAST~\cite{mueller2024sast} & 0.17 & 0.41 & 0.21 & 0.15 & 0.14 & 0.15 \\
%\midrule % Use \midrule to separate main body from the "Ours" row
\rowcolor{blue!20!white} \textbf{OCSD} &  \textbf{0.34}  &  \textbf{0.47} & \textbf{0.45} & \textbf{0.45} & \textbf{0.44} & \textbf{0.44} \\
\bottomrule
\end{tabular}
}
\end{table}

\subsection{Main Results}
We conduct a comprehensive quantitative evaluation against recent state-of-the-art methods on the two datasets. We show our main results on HiK and HOI-M\textsuperscript{3} for short-term forecasting in Tab.~\ref{tab:main} and long-term forecasting in Tab.~\ref{tab:comparisons1}. Our model demonstrates superior performance in both short-term path and pose accuracy and long-term motion realism and diversity, outperforming specialized methods on their respective benchmarks. 
Specifically, our method achieves the best performance in all metrics for HiK in the short-term setting, except for the $0.5$-second path error, where we trail HUMOF by $0.1$ mm. Notably, the longer the time horizon considered, the larger the relative benefit of our method: after one second, we improve over HUMOF by $44.7$ mm, while after two seconds this increases to $121.5$ mm (31.3\%). The pose error results are better for all four time-points considered; however, the margins are narrower. 
On HOI-M\textsuperscript{3}, a similar trend is visible where OCSD outperforms the baselines in all aspects. 

The comparison to long-term pose forecasting models in both realism and diversity further validates the benefits of our approach. We keep a high diversity (UMWR) across the entire ten seconds. This demonstrates that our model captures a much wider distribution of plausible human motion. In contrast, predictions by methods such as SAST tend to ``freeze'' after a certain amount of time, where the people stop moving entirely.
We provide the related examples in the supplementary material.
Furthermore, we achieve state-of-the-art results for the realism metric NDMS, improving over SiMLPe by $25.9\%$, from $0.27$ to $0.34$. This shows that our predicted motion segments are more similar to the actual motion.

\subsection{Ablation studies}
While we discuss the three main ablations regarding object conditioning, social and object context, and loss term below, we provide additional experiments in the supplementary material.

\textbf{Object Conditioning.} We validate the design of our object conditioning mechanism by comparing it with two other variants. 
First, we remove FiLM and use the underlying cross-attention directly to inject the object condition, denoted as ``only Cross-Att".
Second, we remove the cross-attention and instead compute a masked mean-pooling over valid object tokens to obtain only a global object context per person and timeframe, denoted as ``only FiLM". 
From the results in Tab.~\ref{tab:ablation_condition}, we find that FiLM itself provides the weakest form of conditioning. Vanilla cross-attention improves upon FiLM by providing more detailed context, as is visible in the $7.8$ mm improvement in the mean path error. Nonetheless, our proposed object context integration module yields further improvements in the results. Specifically, we increase the mean path error by another $4.2$ mm compared to cross-attention and $12.0$ mm with respect to FiLM. 

\textbf{Social and Object Context.}
Next, we validate the benefits of incorporating social interaction and object information against three ablated versions: ``self + social", which removes the entire object-conditioning pathway, ``self + object", which removes the social encoder and merging layers, and ``only self", which removes both social and object context.

\begin{table*}[t]
\centering
\caption{\textbf{Ablating object integration via cross-attention and FiLM.} We report the path and pose error on HiK. Our combined mechanism is important to get the best results.}
\label{tab:ablation_condition}
\resizebox{0.9\textwidth}{!}{
\begin{tabular}{l rrrrr r rrrrr}
\hline
Method & \multicolumn{5}{c}{Path Error (mm)} & & \multicolumn{5}{c}{Pose Error (mm)} \\
\cline{2-6} \cline{8-12}
& 0.5s & 1.0s & 1.5s & 2.0s & mean & & 0.5s & 1.0s & 1.5s & 2.0s & mean \\
\hline
\rowcolor{blue!20!white}\textbf{OCSD }& \textbf{78.9} & \textbf{132.7} & \textbf{197.8} & \textbf{266.9} & \textbf{137.6} & & \textbf{69.8} & \textbf{99.4} & \textbf{112.3} & 121.9 & \textbf{88.4} \\
only Cross-Att. & 83.0 & 138.5 & 204.5 & 268.8 & 141.8 & &  70.3 & 100.2 & 113.3 &  \textbf{121.6} &  88.9 \\
only FiLM  & 87.5 & 146.3 & 213.9 & 284.7 & 149.6 & & 74.5 & 102.3 & 114.3 & 123.9 & 91.6 \\
\hline
\end{tabular}
}
\end{table*}

\begin{table*}[t]
\centering
\caption{\textbf{The importance of context for short-term forecasting.} We show the path and pose error on HiK. Integrating all forms of context leads to the best performance.} 
\label{tab:ablation_context_short}
\resizebox{0.9\textwidth}{!}{
\begin{tabular}{l rrrrr r rrrrr}
\hline
Method & \multicolumn{5}{c}{Path Error (mm)} & & \multicolumn{5}{c}{Pose Error (mm)} \\
\cline{2-6} \cline{8-12}
& 0.5s & 1.0s & 1.5s & 2.0s & mean & & 0.5s & 1.0s & 1.5s & 2.0s & mean \\
\hline
\rowcolor{blue!20!white}\textbf{OCSD }& \textbf{78.9} & \textbf{132.7} & \textbf{197.8} & \textbf{266.9} & \textbf{137.6} & & 69.8 & 99.4 & 112.3 & \textbf{121.9} & \textbf{88.4} \\
self + social  & 81.6 & 141.9 & 213.8 & 281.5 & 146.5 & &\textbf{69.7} & 100.3 & 112.9 &  124.3 &  89.1 \\
self + object & 83.1 & 140.1 & 208.1 & 277.4 & 144.7 & & 72.5 & 102.7 & 114.9 & 123.5 & 90.4 \\
only self  & 84.5 & 146.2	& 217.7 & 289.8 & 149.9 & & 70.8 & \textbf{99.2} & \textbf{112.2}  & 122.9 & 88.5 \\
\hline
\end{tabular}
}
\end{table*}

\begin{table}[t]
\centering
\caption{\textbf{The impact of context for long-term forecasting.} On the HiK dataset, realism score (NDMS) increases with the context while diversity score (UMWR) is lower because of the introduced constraints from environment and people.} 
\label{tab:ablation_context_long}
\resizebox{0.6\textwidth}{!}
{
\begin{tabular}{l c  ccccc}
\toprule
Method & {NDMS $\uparrow$} &  \multicolumn{5}{c}{UMWR $\uparrow$} \\
\cmidrule(l){3-7} % Creates a partial rule under the UMWR columns
& & 2s & 4s & 6s & 8s & 10s \\
\midrule
\rowcolor{blue!20!white}\textbf{OCSD} &   \textbf{0.34}  &  0.47 & 0.45 & \textbf{0.45} & 0.44 & 0.44   \\
self + social  & 0.33 & 0.47  & 0.45  & 0.44 & 0.44 & 0.44 \\
self + object &  0.33 & 0.47 & 0.45   & \textbf{0.45} & 0.44 & 0.44 \\
only self  & 0.32 & \textbf{0.48}  & \textbf{0.46}  & \textbf{0.45} & \textbf{0.45} & \textbf{0.45} \\
\bottomrule
\end{tabular}
}
\end{table}

For short-term forecasting, Tab.~\ref{tab:ablation_context_short} shows the effect of removing object and social information. For path error, both context are clearly beneficial. Incorporating social context reduces mean path error by $3.4$ mm, while object information reduces it by $5.2$ mm. By combining both forms, our full model achieves the best result overall, lowering the mean path error by $12.3$ mm. This result demonstrates that the two sources of context are complementary rather than redundant, which is intuitive since the global trajectory is affected by where other people and objects are. In contrast, local pose in short horizons is primarily determined by the person's own dynamics, so the ``only self" model already achieves a low pose error. Therefore, the impact of context is not obvious in the mean over different types of activities. We provide more details on the context contribution in the supplementary material by reporting the metrics for representative activities.

For long-term forecasting, Tab.~\ref{tab:ablation_context_long} shows that the NDMS realism metric also improves with more context.
Interestingly, the diversity of samples as measured by the UMWR metric is slightly higher without any context, as more movements are plausible without any constraints from the environment. 
Nonetheless, the diversity of the different variants is similar.

\begin{table*}[t]
\centering
\caption{\textbf{Ablating loss weighting.} We report the average path and pose error over four samples on HiK. Using the loss weighting improves the results.}
\label{tab:ablation_loss}
\resizebox{0.9\textwidth}{!}
{
\begin{tabular}{ll rrrrr rrrrr}
\hline
 Method & \multicolumn{5}{c}{Path Error (mm)} & &\multicolumn{5}{c}{Pose Error (mm)} \\
\cline{2-6} \cline{8-12}
 & 0.5s & 1.0s & 1.5s & 2.0s & mean & & 0.5s & 1.0s & 1.5s & 2.0s & mean \\
\hline 
\rowcolor{blue!20!white}\textbf{OCSD}& \textbf{78.9} & \textbf{132.7} & \textbf{197.8} & \textbf{266.9} & \textbf{137.6} & & 69.8 & \textbf{99.4} & \textbf{112.3} & \textbf{121.9} & \textbf{88.4} \\
w/o weighting  & 82.1 & 139.8 & 208.2 & 277.4 & 144.8 & &\textbf{69.6} & 101.6 & 117.3 & 124.8 & 90.0 \\
\hline
\end{tabular}
}
\end{table*}

\begin{figure*}[t]
  \centering
  \setlength\tabcolsep{1pt}
  \begin{tabular}{c}
    \includegraphics[width=0.9\linewidth]{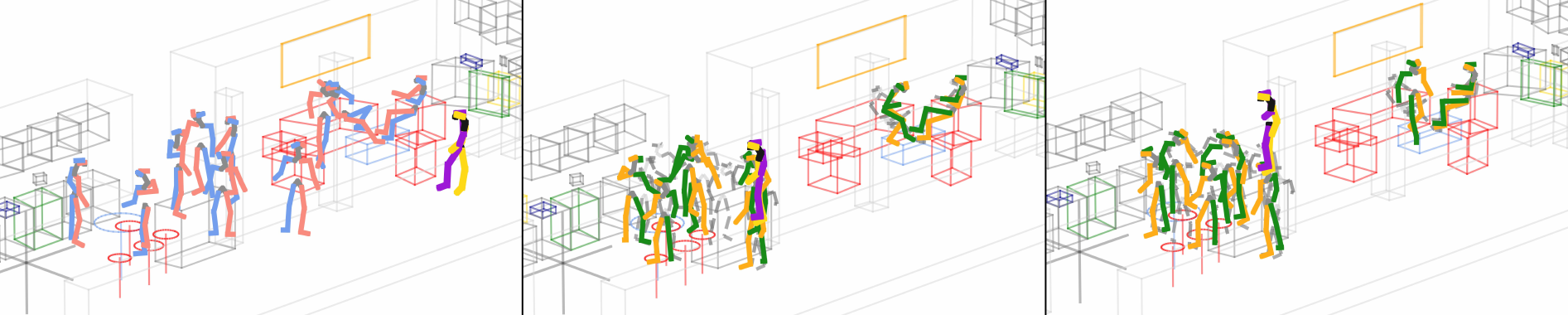} \\
  \end{tabular}
    \caption{\textbf{Qualitative examples of 10-second predictions for the HiK dataset.} 
    We visualize two generated samples for the same input.
    We show the last observed frame on the left, followed by the final predicted frame from two distinct generated samples.
    These examples demonstrate the model's ability to generate diverse predictions, as shown by the purple/yellow person walking to different locations.}
    \label{fig:sec2hik3-main}
\end{figure*}
\begin{figure*}[t]
  \centering
  \setlength\tabcolsep{0pt}
  \begin{tabular}{ccc}
    \includegraphics[width=0.33\linewidth]{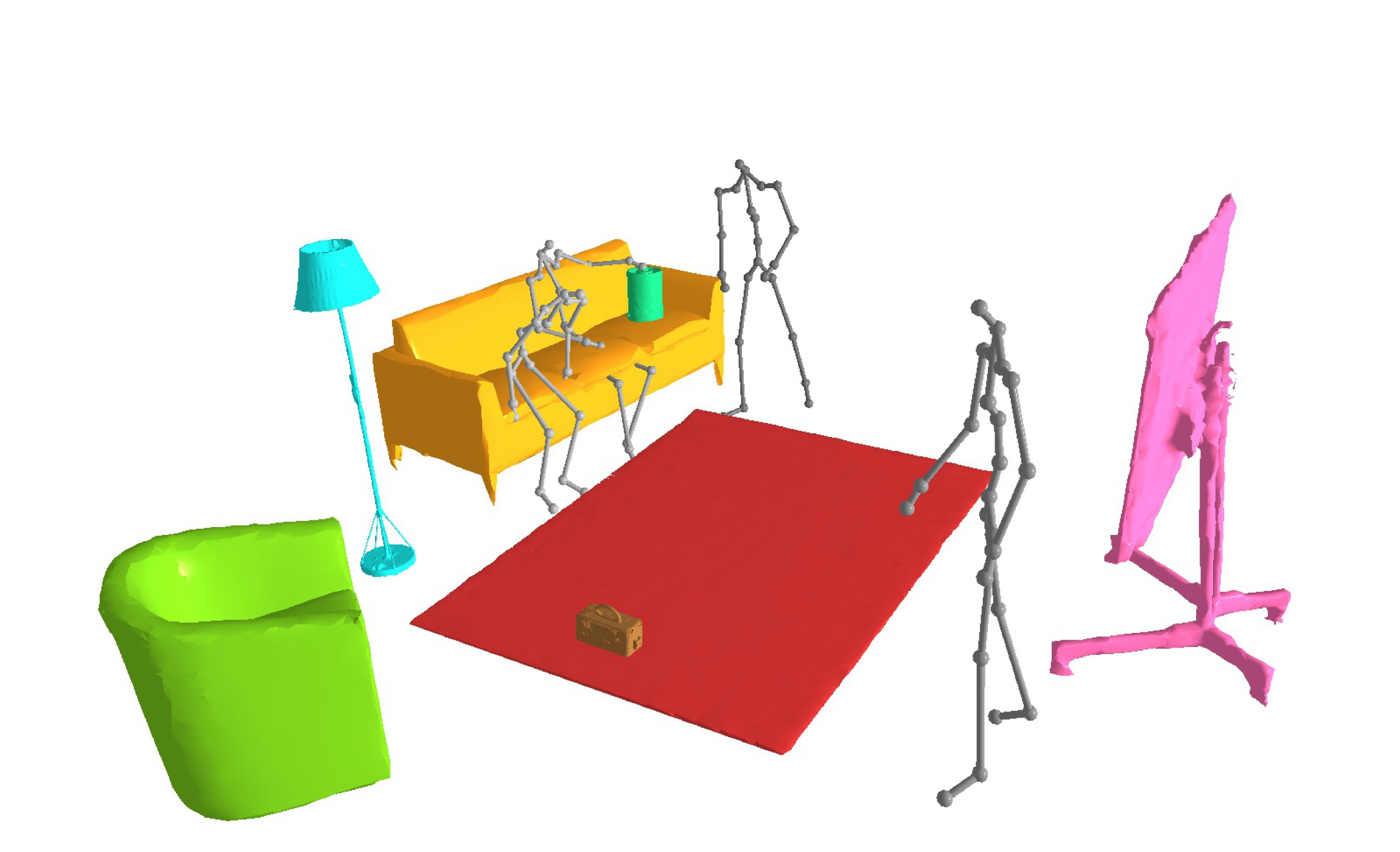} &
    \includegraphics[width=0.33\linewidth]{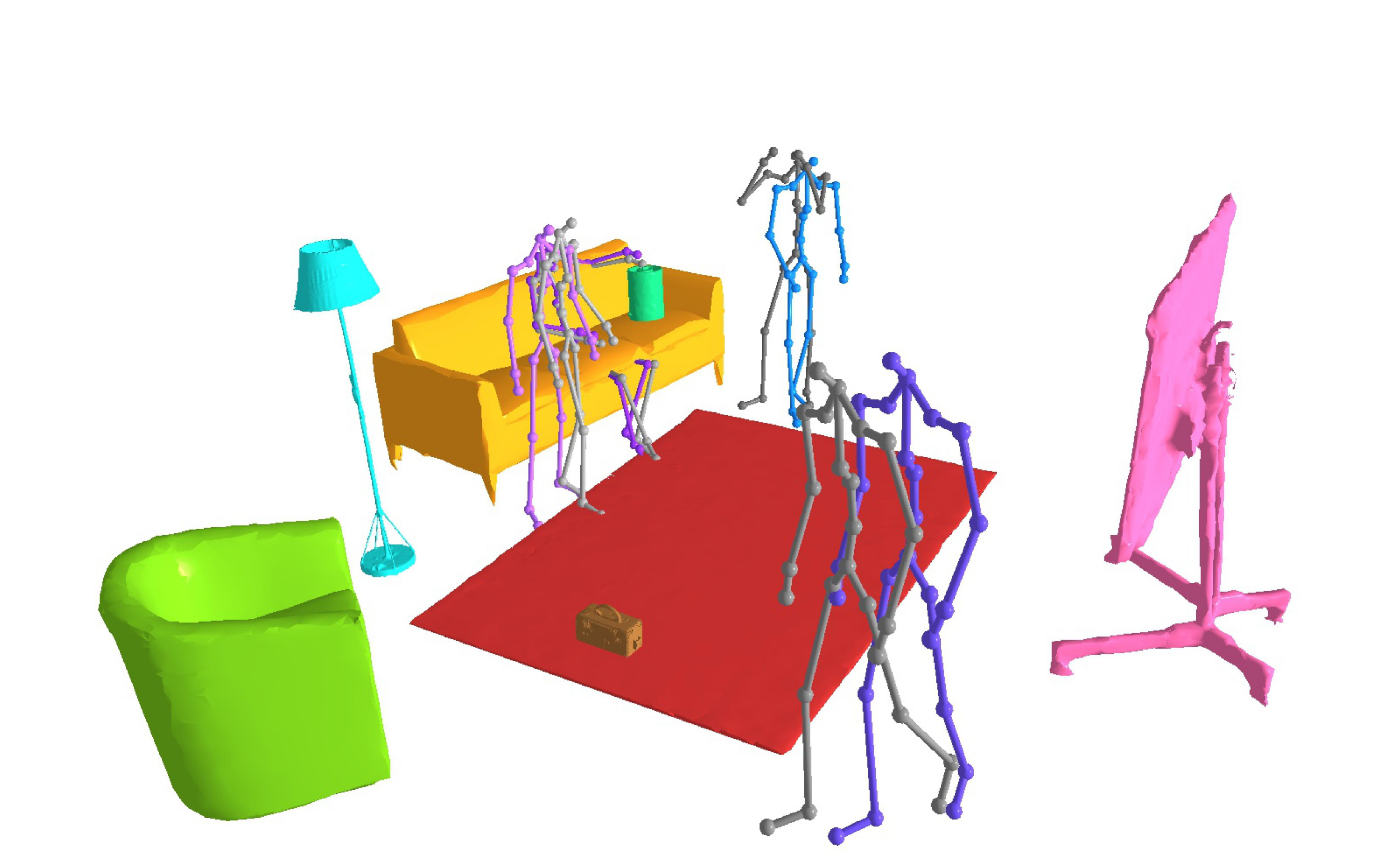} &
    \includegraphics[width=0.33\linewidth]{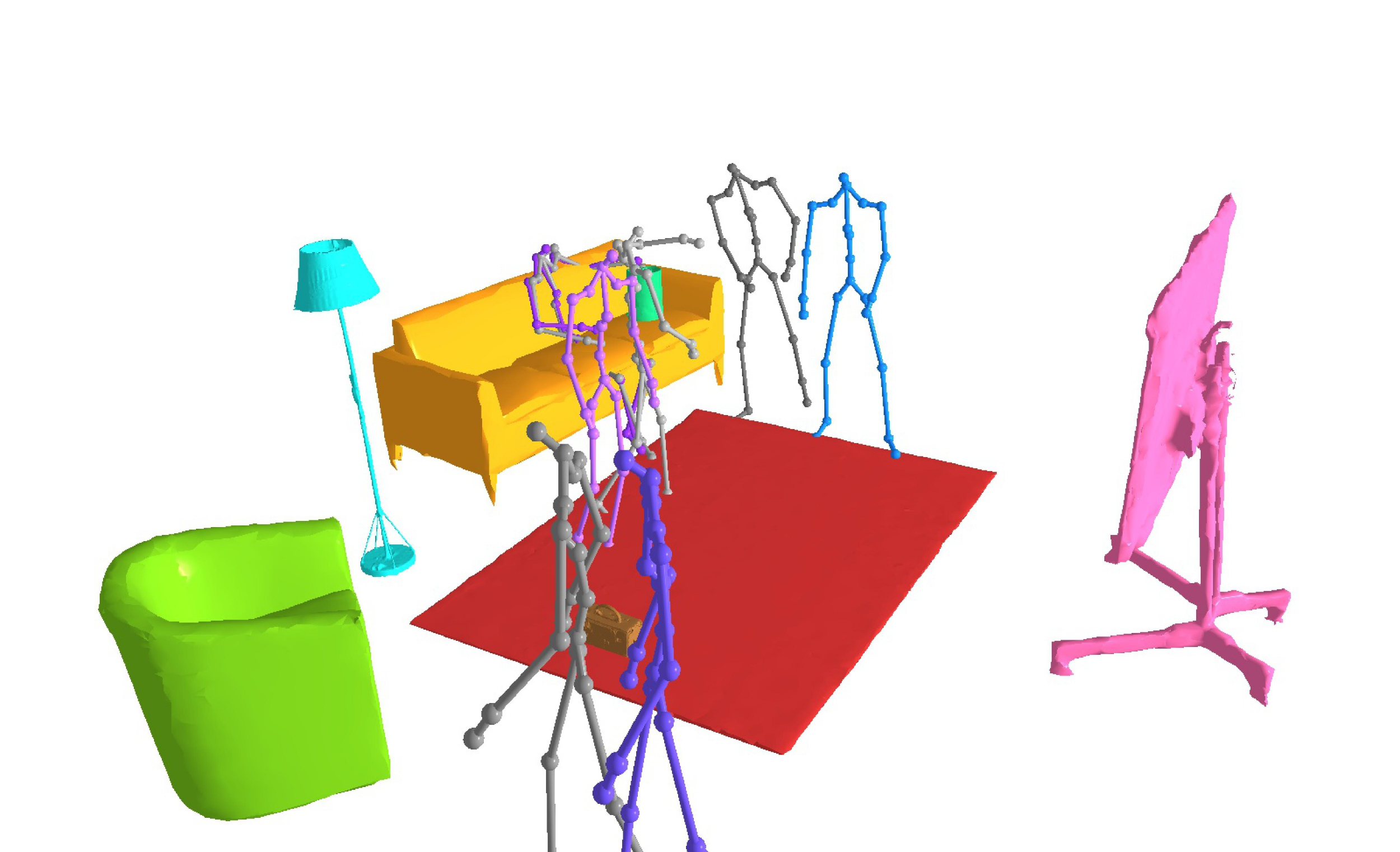} \\
  \end{tabular}
    \caption{\textbf{2-second prediction for the HOI-M\textsuperscript{3} dataset.}  We show the last observed frame on the left. The next two panels display forecast poses after 1 second (middle) and 2 seconds (right).
    Purple skeletons represent our predictions, while gray skeletons represent ground-truth poses. The model correctly continues the stand up motion from the sofa and the walk motion of the other person.}
    \label{fig:sec2hoi2}
\end{figure*}

\textbf{Loss Aggregation.}
Finally, we ablate the way the diffusion loss is aggregated over channels. We compare to a baseline where we remove the component-wise decomposition of Eq.~\eqref{eq:loss}. Instead, we compute a single scalar loss as the mean of the prediction error over all elements.
The results in Tab.~\ref{tab:ablation_loss} show that our component-wise decomposition, which assigns more weight to the global positioning of the sequences, is important to reach a good performance.

\subsection{Qualitative Results}
\label{sec:quali}
Beyond quantitative metrics, we provide qualitative visualizations to demonstrate the fidelity, realism, and diversity of our generated motions. 
Specifically, a key advantage of our diffusion framework is the ability to model a distribution of plausible futures. Fig.~\ref{fig:sec2hik3-main} illustrates this by showing two distinct 10-second predictions sampled from the same input.
This capability to generate a wide range of realistic behaviors directly explains our model's significantly higher UMWR (diversity) score compared to all baselines, which tend to produce only a single, deterministic-looking outcome.
In Fig.~\ref{fig:sec2hoi2} from HOI-M\textsuperscript{3}, the person sitting on the sofa stands up correctly and continues moving, the walking person continues walking smoothly, and the standing person achieves the same pose as the ground truth at a slightly different but reasonable location.
We provide additional qualitative results and discuss limitations in the supplementary material. 
%Furthermore, we include Sec.~\ref{sec:limits} for limitations of the model.  

\section{Conclusion}
\label{sec:conc}
We tackled the challenging problem of forecasting multi-person human motion in complex scenes. We presented OCSD, a model that integrates individual dynamics, social interactions, and object influence into a unified generative framework. By modulating the denoising process with object cues at every timestep and integrating per-person social context, OCSD captures fine-grained relations while maintaining flexibility in group size, interaction complexity, and the diversity of possible futures.
Our experiments on the HiK and HOI-M\textsuperscript{3} benchmarks demonstrate substantial improvements over existing methods. Overall, our results show that holistic approaches like OCSD, which integrate both social and object information, are necessary for accurate motion forecasting in complex scenes.

\section*{Acknowledgements}
The work has been supported by the ERC Consolidator Grant FORHUE (101044724).
%For the computations involved in this research, we acknowledge EuroHPC Joint Undertaking for awarding us access to Leonardo at CINECA, Italy, through EuroHPC Regular Access Call - proposal No. EHPC-REG-2025R01-218.

% Note: if you want to use up all of the allowed space for the paper,
%       the bibliography will start on top of page 13. Furthermore,
%       from page 13 onwards, there will be *only* bibliography, no more
%       figures/tables.
\bibliographystyle{splncs04}
\bibliography{egbib}

%\input{X_suppl}
\input{088-supp}

\end{document}

%% file: 088-supp.tex
%%%%%%%%%%%%%%%%%%%%% Add submission id, track, and title. %%%%%%%%%%%%%%%%%%%%%

%\maketitle              % typeset the header of the contribution

%\input{0_abstract}    
%\input{1_intro}
%\input{2_related}
%\input{3_method}
%\input{4_exp}
%\input{5_conc}

%\section*{Acknowledgements}
%The work has been supported by the Federal Ministry of Research, Technology and Space (BMFTR) under grant no. 16IS22094A WEST-AI and the ERC Consolidator Grant FORHUE (101044724).
%For the computations involved in this research, we acknowledge EuroHPC Joint Undertaking for awarding us access to Leonardo at CINECA, Italy, through EuroHPC Regular Access Call - proposal No. EHPC-REG-2025R01-218.

% Note: if you want to use up all of the allowed space for the paper,
%       the bibliography will start on top of page 13. Furthermore,
%       from page 13 onwards, there will be *only* bibliography, no more
%       figures/tables.

\appendix
\clearpage
\setcounter{table}{0}
\setcounter{figure}{0}
\setcounter{equation}{0}

\renewcommand{\theHsection}{Appendix.\thesection} 
\renewcommand{\thefigure}{S\arabic{figure}}
\renewcommand{\thetable}{S\arabic{table}}
\renewcommand{\theequation}{S\arabic{equation}}

\begin{center}
%Multi-Person Human Motion Forecasting in Complex Scenes\\
%    {\Large \textbf{Supplementary Material}}
{\Large \textbf{Multi-Person Human Motion Forecasting in Complex Scenes}}\\
{Supplementary Material}
\end{center}

\section{Datasets}
\label{sec:extdatasets}

Here, we provide the full details on the datasets used in our paper:

\begin{itemize}
    \item \textbf{Humans in Kitchens~\cite{tanke2023humans}.}
    HIK is a motion-capture dataset of people interacting in four kitchen environments with minimal scripting and rich everyday activities. Each recording contains a dynamic number of people with a maximum of 16, and 50 annotated objects visible at the same time, providing complex social interactions and dense scene context. Following the standard protocol, we train on three kitchens (A, B, and C) and evaluate on curated sequences from kitchen D, with a focus on transitional moments. 
    \item \textbf{HOI-M\textsuperscript{3}~\cite{zhang2024hoi}.} 
    HOI-M\textsuperscript{3} is a large-scale multi-human multi-object dataset captured in realistic indoor rooms. It contains 199 long sequences with up to 5 people and at least 5 objects per scene, and all humans and objects are tracked in 3D. We use object vertices as object representation and 49 livingroom sequences publicly available (1-51, except 43 and 45). We set aside $20\%$ as the test set following~\cite{sun2025humof}. 
\end{itemize}

For the HiK dataset, we adhere to the standard evaluation protocol, utilizing the official test split and evaluation script released by~\cite{tanke2023humans,mueller2024sast}. For the HOI-M\textsuperscript{3} dataset, we evaluated performance on a test set split by using the data processing pipeline of~\cite{sun2025humof}. 
During inference, we predicted future motions for all frames in these sequences using a sliding window approach with a stride of $1$ second.

\section{Metrics}
\label{sec:extmetrics}

The metrics used in our work are as follows:
\begin{itemize}
 \item \textbf{Path Error.}
This metric evaluates the accuracy of the predicted global trajectory. We define it as the Euclidean distance ($L_2$ norm) between the predicted 3D position of the root joint ($\hat{p}_{root}$) and its corresponding ground-truth position ($p_{root}$). For a future time step $t$, the path error is:
\begin{equation}
    \mathcal{E}_{path}(t) = || \hat{p}_{root}^{t} - p_{root}^{t} ||_2
\end{equation}
\item \textbf{Pose Error.} This metric evaluates the accuracy of the predicted local body configuration, independent of global translation. We use the standard Mean Per Joint Position Error (MPJPE), which is the mean Euclidean distance across all $J-1$ non-root joints after aligning the predicted and ground-truth poses at the root. For a future time step $t$, the pose error is:
\begin{equation}
    \mathcal{E}_{pose}(t) = \frac{1}{J-1} \sum_{j=1}^{J-1} || (\hat{p}_{j}^{t} - \hat{p}_{root}^{t}) - (p_{j}^{t} - p_{root}^{t}) ||_2
\end{equation}
    where $\hat{p}_{j}^{t}$ and $p_{j}^{t}$ are the predicted and ground-truth 3D positions of the $j$-th joint. 
 \item \textbf{Normalized Directional Motion Similarity (NDMS).}
For long-term and stochastic forecasting, per-frame accuracy metrics are insufficient due to the fact that they penalize plausible predictions that deviate from the single ground truth sequence. Therefore, we evaluate the realism of generated motion using the Normalized Directional Motion Similarity (NDMS) score \cite{tanke2021intention}, which compares generated motion against a distribution of real-world examples.  
To evaluate a predicted motion sequence $\mathbf{\chi}$, NDMS divides it into overlapping short segments, referred to as motion words. For each motion word, which is $8$ frames long, the metric calculates a similarity score by finding its Nearest Neighbor (NN) in a reference database ($D$) composed of real motion snippets. This use of a reference database and NN search allows the measure to account for the multi-modal nature of human movement. The core similarity function $\Psi$ compares two motion words ($\mathbf{x}$ and $\mathbf{y}$) by evaluating two key factors for each joint $j$ at time $t$:
\begin{enumerate}
\item  Directional Similarity: The alignment of the joint velocity vectors ($\mathbf{\dot{x}}_{t,j}$ and $\mathbf{\dot{y}}_{t,j}$).
\item Magnitude Ratio: The ratio of the velocity magnitudes, ensuring similar speeds.
\end{enumerate}
The resulting NDMS score provides a plausibility measure between $0$ and $1$, where higher values indicate more realistic motion. NDMS penalizes discontinuities which are highly visible to human observers, such as abrupt transitions between the observed and forecast motion.
 \item \textbf{Unique Motion Word Ratio (UMWR).}
 A good stochastic model should not only produce realistic motions but also avoid generating the same plausible motion repeatedly. For this, we use the Unique Motion Word Ratio (UMWR) introduced in \cite{mueller2024sast}. The UMWR calculation leverages the same underlying mechanism as NDMS. It uses the same reference database $D$, composed of motion words of $8$ frames. For a single-person predicted sequence $\mathbf{\chi}$, the UMWR is calculated as the ratio of unique Nearest Neighbors (NN) found in the reference set $D$ to the total number of motion words in the sequence. Formally, the metric is defined in \cite{mueller2024sast} as
 \begin{equation}
 \text{UMWR}(\mathbf{\chi}) =\frac{|\{\text{NN}(\mathbf{\chi}_{1:\kappa}), \text{NN}(\mathbf{\chi}_{2:\kappa+1}), \dots \text{NN}(\mathbf{\chi}_{|\mathbf{\chi}|+1-\kappa:|\mathbf{\chi}|) }\}|}{|\mathbf{\chi}|+ 1 - \kappa}
 \end{equation}
 where  $\kappa=8$ is the length of the motion words and $\chi$ predicted motion of the selected person with last observed frame.   
A score closer to $1.0$ signifies high diversity, as the model generates a wide variety of movements that map to many different real motion words. A low score indicates repetitive or frozen motion, where the entire sequence maps to only a few unique motion words from $\mathcal{D}$.
\end{itemize}

\section{Additional Implementation Details}
\label{sec:extimplement}
We trained all models on a single RTX-$4090$ GPU. To accommodate memory constraints during training, we adjusted the batch size according to the prediction horizon: we used a batch size of $64$ for short-term ($2$-second) prediction models and a batch size of $16$ for long-term ($10$-second) prediction models, where we also adjust the learning rate to $1 \times 10^{-4}$ to accommodate the reduced batch size. Moreover, in the long-term setting, each temporal feature sequence at the bottleneck ($h_{1}$) is passed through multi-head self-attention with pre-norm and a residual connection to capture long-range temporal dependencies within each local context. Each residual block uses two successive dilated temporal convolutions (d=2 and d=4) instead of unit dilation to extend the temporal receptive field without increasing parameter count.
We employ a dynamic sampling strategy. Rather than using fixed, strided subsequences, each training sample is a clip randomly selected from a valid start position within any sequence.  We use 1000 denoising steps.
For inference on the HOI-M\textsuperscript{3} dataset, our approach requires as SAST 2s. To put this into context, HUMOF reports 43ms.

\section{Additional Qualitative Results}
In this section, we show further qualitative results of our model for both datasets. We use SAST \cite{mueller2024sast} for comparison, as the code for HUMOF~\cite{sun2025humof} was not publicly available at the time of preparing the manuscript. 
\label{sec:extqual}
\subsection{HiK dataset: 2-second prediction}
\begin{figure*}[t]
  \centering
  \setlength\tabcolsep{1pt}
  \begin{tabular}{ccc}
    \includegraphics[width=0.37\linewidth]{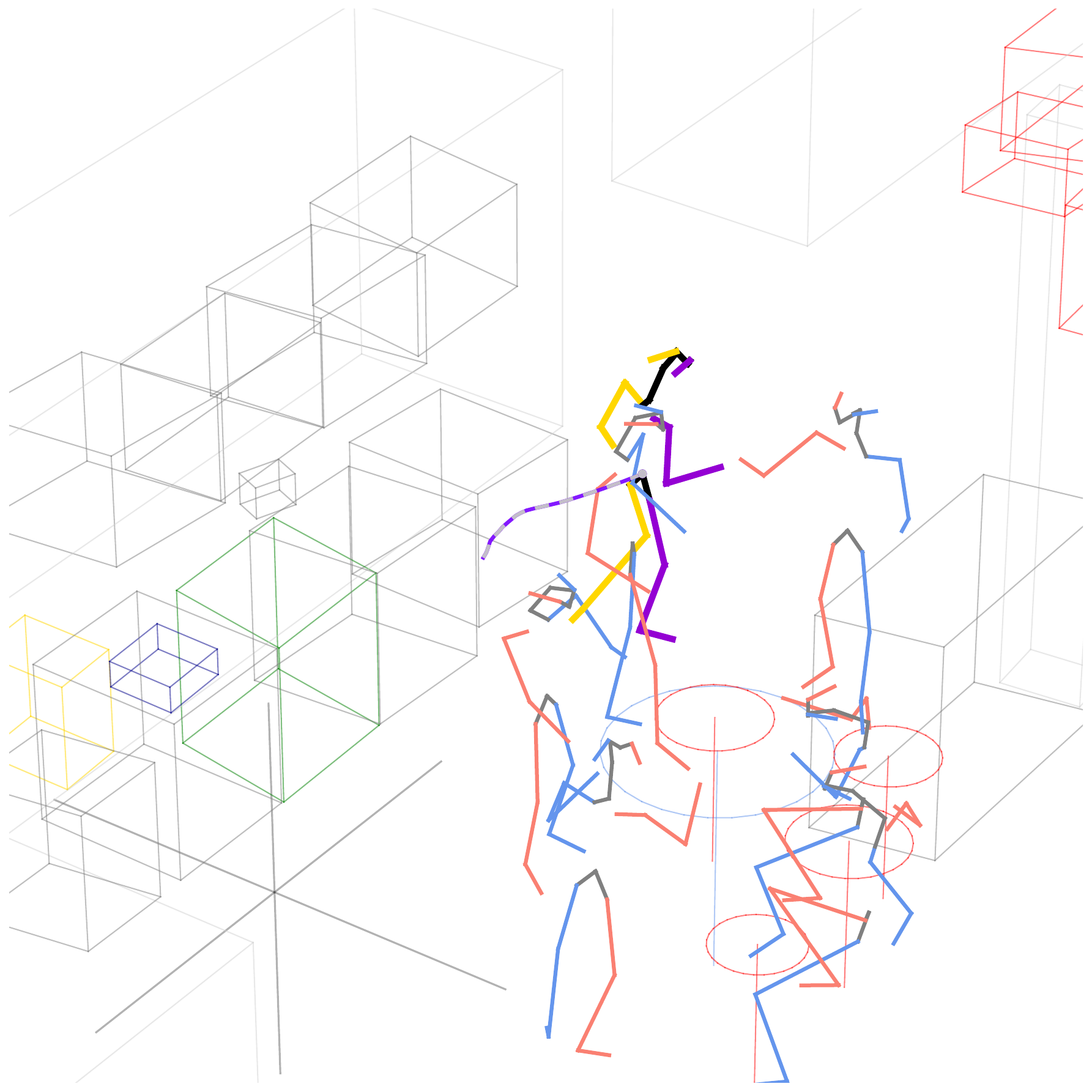} &
    \includegraphics[width=0.37\linewidth]{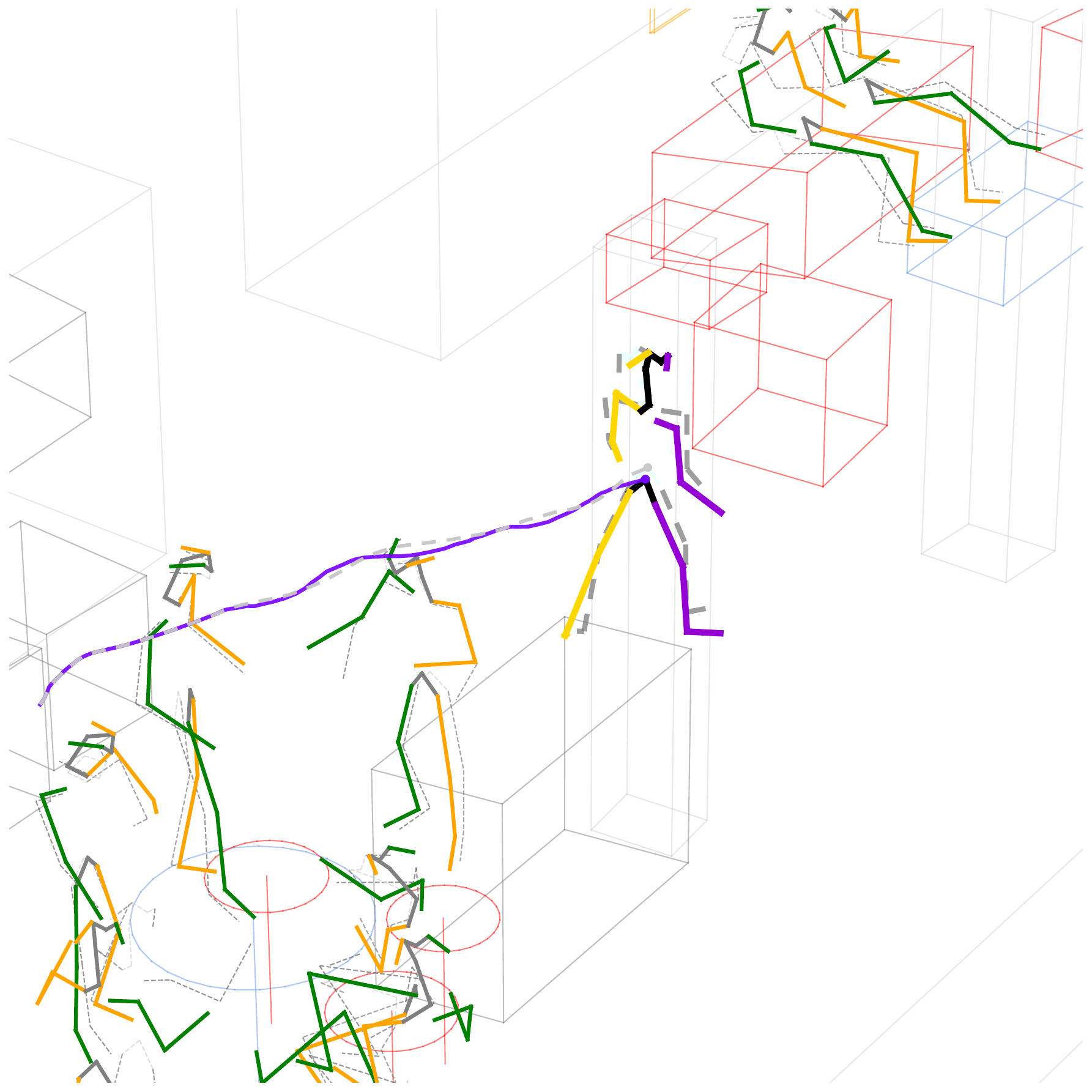} &
    \includegraphics[width=0.17\linewidth]{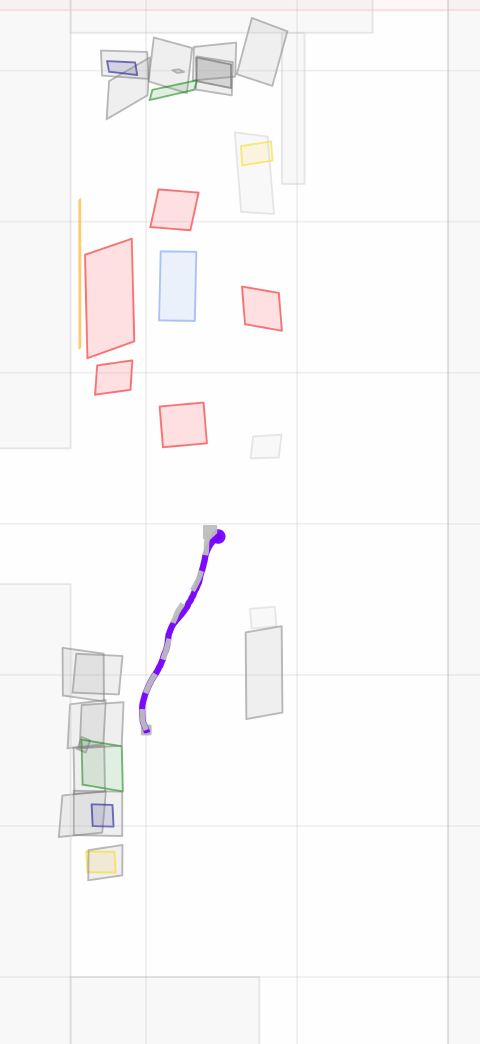} \\
    (a) Frame 24 ($t=1s$) & (b) Frame 74 ($t=3s$) & (c) Bird's-eye  \\
  \end{tabular}
    \caption{\textbf{Qualitative comparison on the HiK dataset (2-second prediction).} 
        We compare our predictions to the ground truth (GT).
        %\textbf{Color Legend:} 
        \textbf{Red/Blue:} Past GT poses.
        \textbf{Green/Yellow:} Predicted future poses.
        \textbf{Gray:} Future GT for comparison.
        \textbf{Purple/Yellow:} Predicted future and past GT poses of a person with large motion.
        The long gray and purple lines show the GT and predicted trajectory, respectively.
        OCSD's predictions overlap significantly with the GT (Gray), indicating high accuracy.
        }
    \label{fig:sec2hik1}
\end{figure*}

\begin{figure*}[ht!]
  \centering
  \setlength\tabcolsep{1pt}
  \begin{tabular}{ccc}
    \includegraphics[width=0.32\linewidth]{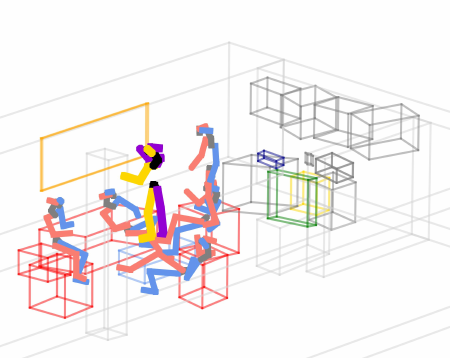} &
    \includegraphics[width=0.32\linewidth]{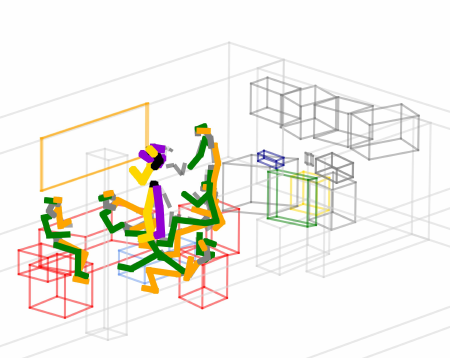} &
    \includegraphics[width=0.32\linewidth]{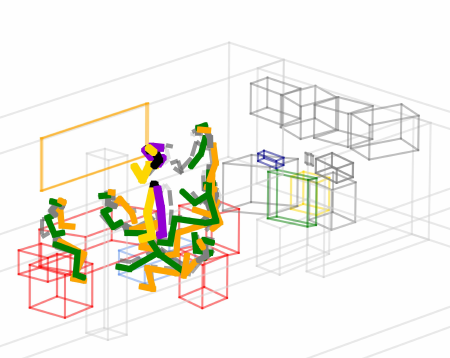} \\
     & (a) whiteboard &  \\
    \includegraphics[width=0.32\linewidth]{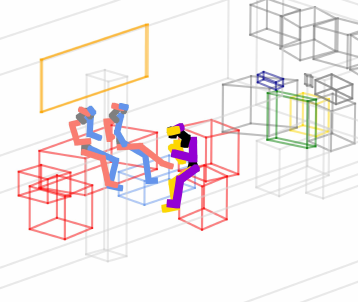} &
    \includegraphics[width=0.32\linewidth]{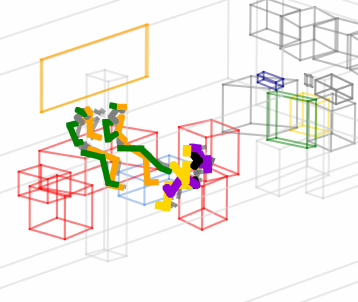} &
    \includegraphics[width=0.32\linewidth]{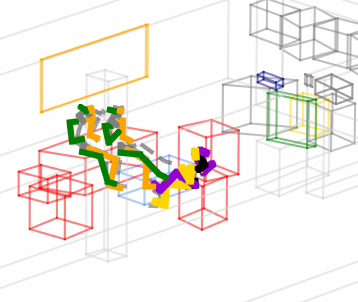} \\
     & (b) sitting down &  \\
  \end{tabular}
    \caption{\textbf{Object interactions on the HiK dataset (2-second prediction).} Each row shows the last observed frame (Frame 24), 1-second prediction (Frame 49), and 2-second prediction (Frame 74). Visualization scheme follows Fig.~\ref{fig:sec2hik1}. Row (a) shows the writing activity with interaction with the whiteboard, while row (b) shows the sitting down activity with interaction with the armchair.}
    \label{fig:sec1hik0}
\end{figure*}

In Fig.~\ref{fig:sec2hik1}, we visualize the predicted motions against ground truth (GT) for a $2$-second prediction window. Our model simultaneously predicts the future motions of all individuals in the scene. As demonstrated in the figure, our method generates highly accurate future poses. Even for the purple/yellow skeleton, which moves most in the scene, the predicted motion aligns closely with the ground truth (gray skeleton), making them difficult to distinguish even at frame 74. This validates the model's ability to predict future motion precisely. In addition, examples for the whiteboard and armchair activities are provided in Fig.~\ref{fig:sec1hik0}.

\begin{figure*}[ht!]
  \centering
  \setlength\tabcolsep{1pt}
  \begin{tabular}{ccc}
    \includegraphics[width=0.32\linewidth]{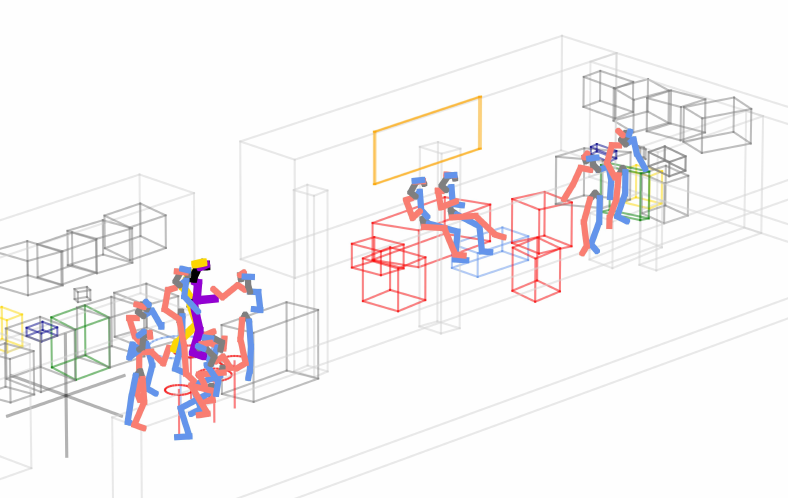} &
    \includegraphics[width=0.32\linewidth]{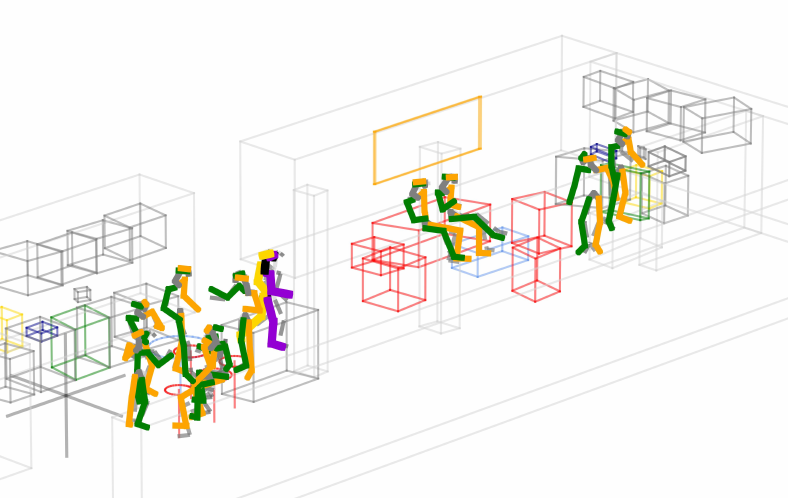} &
    \includegraphics[width=0.32\linewidth]{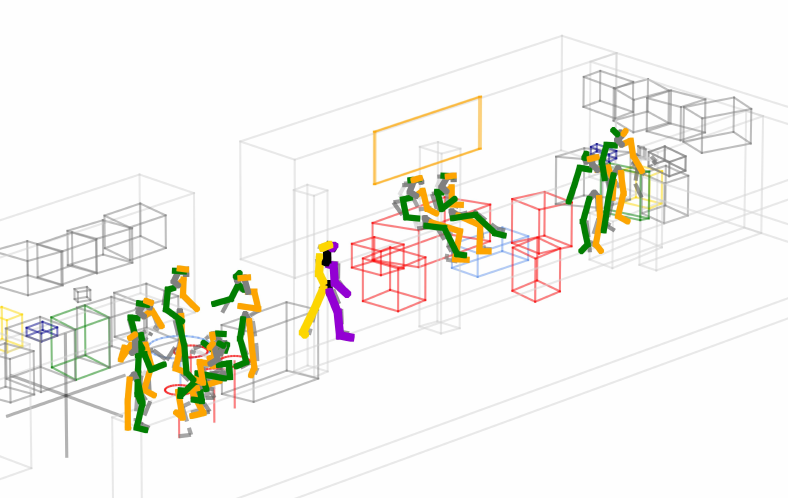} \\
     & (a) OCSD (ours) &  \\
    \includegraphics[width=0.32\linewidth]{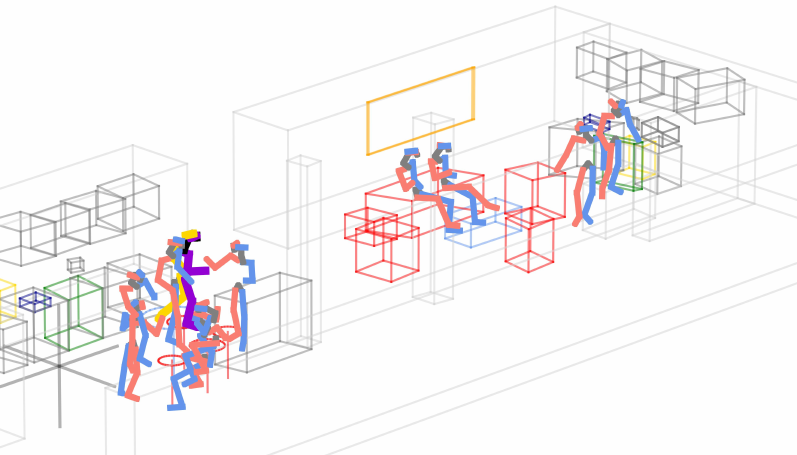} &
    \includegraphics[width=0.32\linewidth]{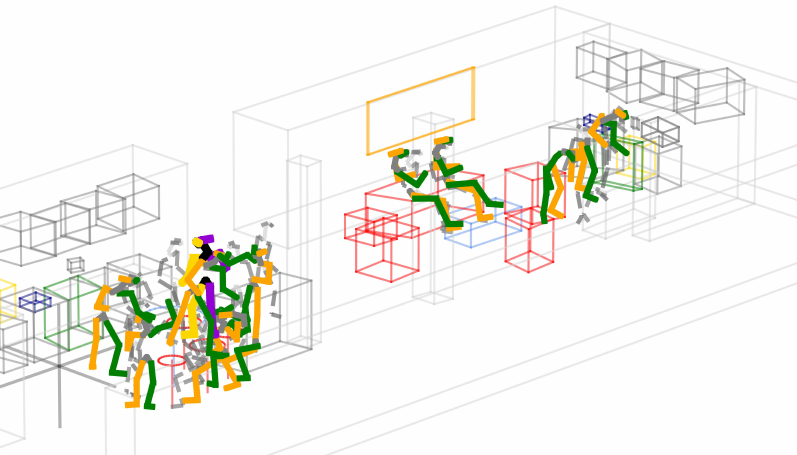} &
    \includegraphics[width=0.32\linewidth]{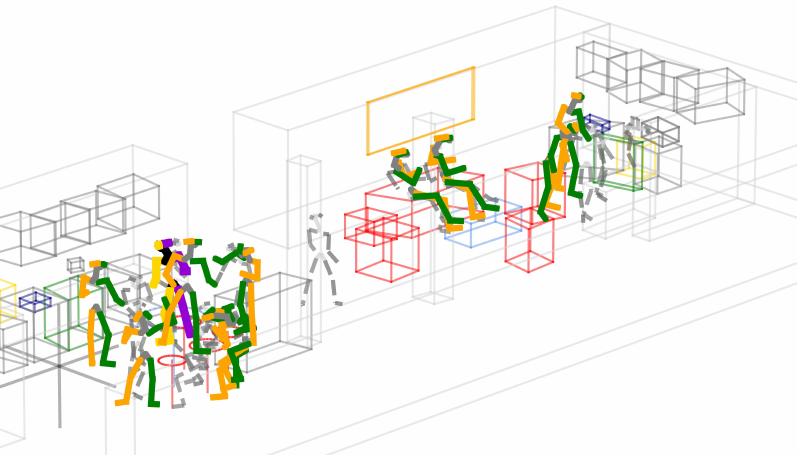} \\
     & (b) SAST &  \\
  \end{tabular}
    \caption{\textbf{Qualitative comparison with SAST (HiK dataset, 2-second prediction).}
    Each row shows the last observed frame (Frame 24), $1$-second prediction (Frame 49), and $2$-second prediction (Frame 74). Visualization scheme follows Fig.~\ref{fig:sec2hik1}. SAST (row b) does not forecast that the purple person walks from the left group to the right group, while OCSD (row a) successfully predicts the walking motion.}
    \label{fig:sec2hik2}
\end{figure*}

\textbf{Comparison with SAST (2-second).} 
In Fig.~\ref{fig:sec2hik2}, we compare our method (OCSD) against the SAST baseline. Both models are trained to predict a $2$-second horizon given a $1$-second input. While our model correctly infers the walking trajectory of the purple/yellow person, SAST predicts that the person will stay close to the last observed position. 

\subsection{HOI-M\textsuperscript{3} dataset: 2-second prediction}
\begin{figure*}[t]
  \centering
  \setlength\tabcolsep{1pt}
  \begin{tabular}{ccc}
    \includegraphics[width=0.32\linewidth]{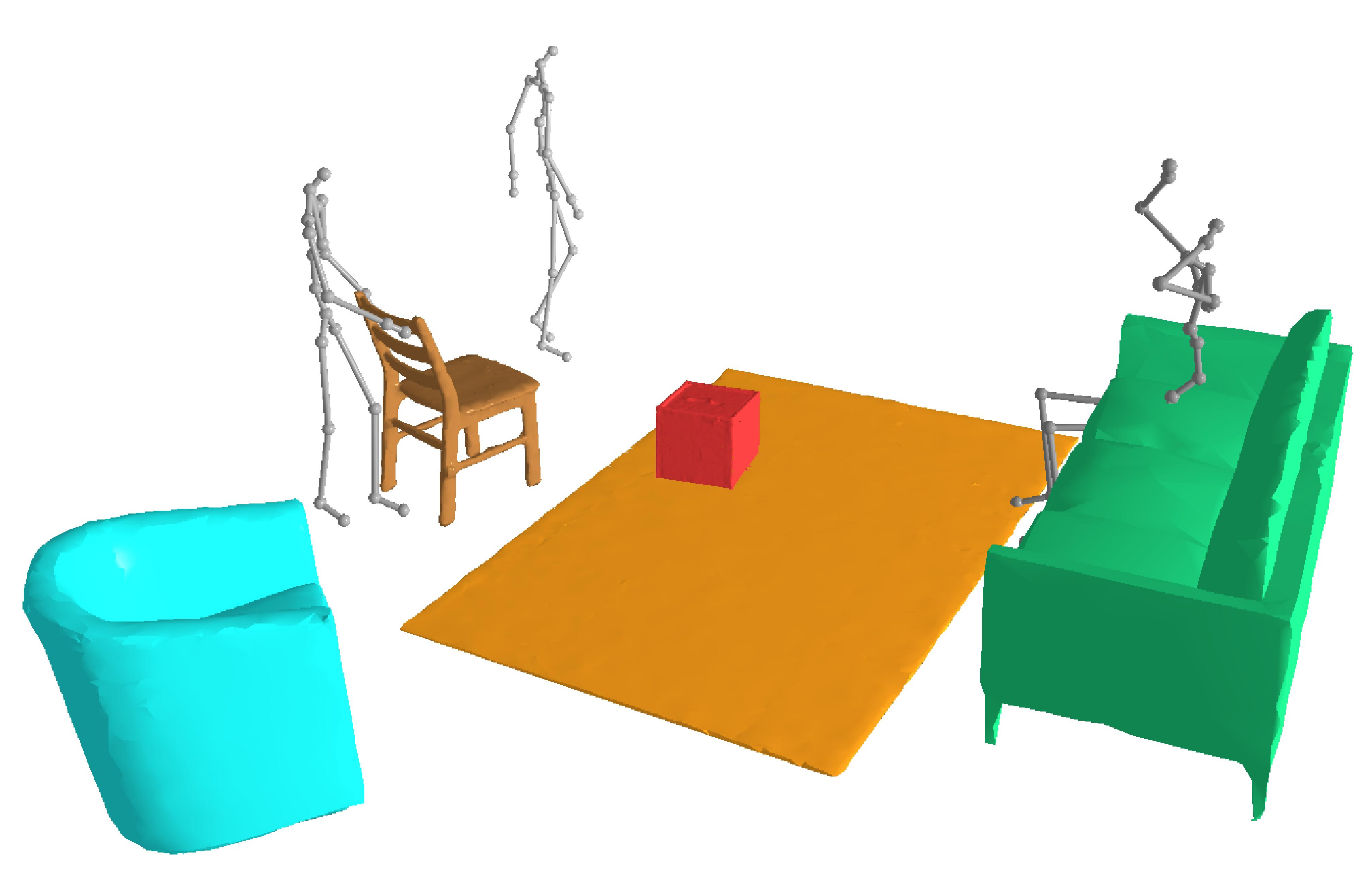} &
    \includegraphics[width=0.32\linewidth]{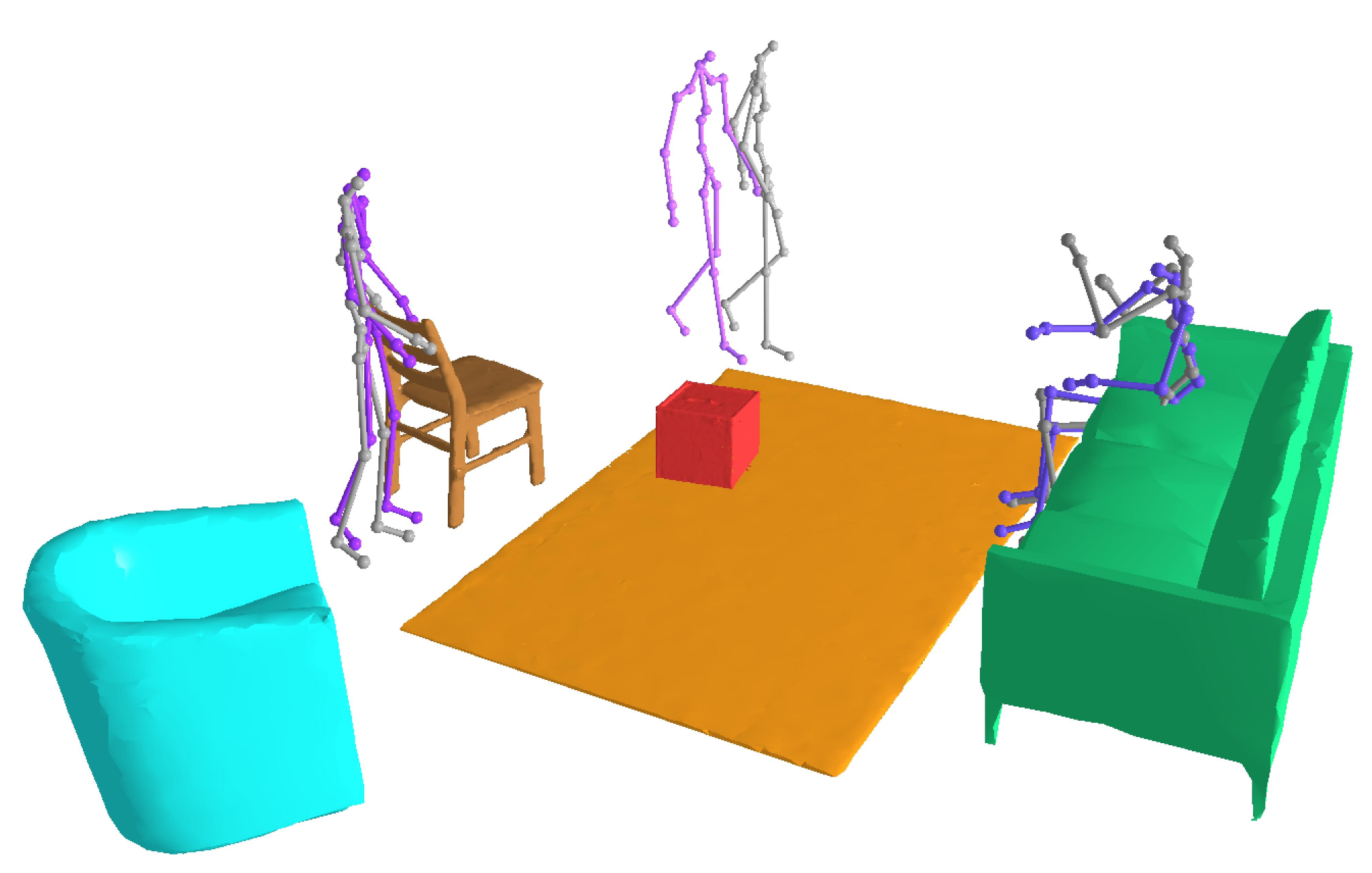} &
    \includegraphics[width=0.32\linewidth]{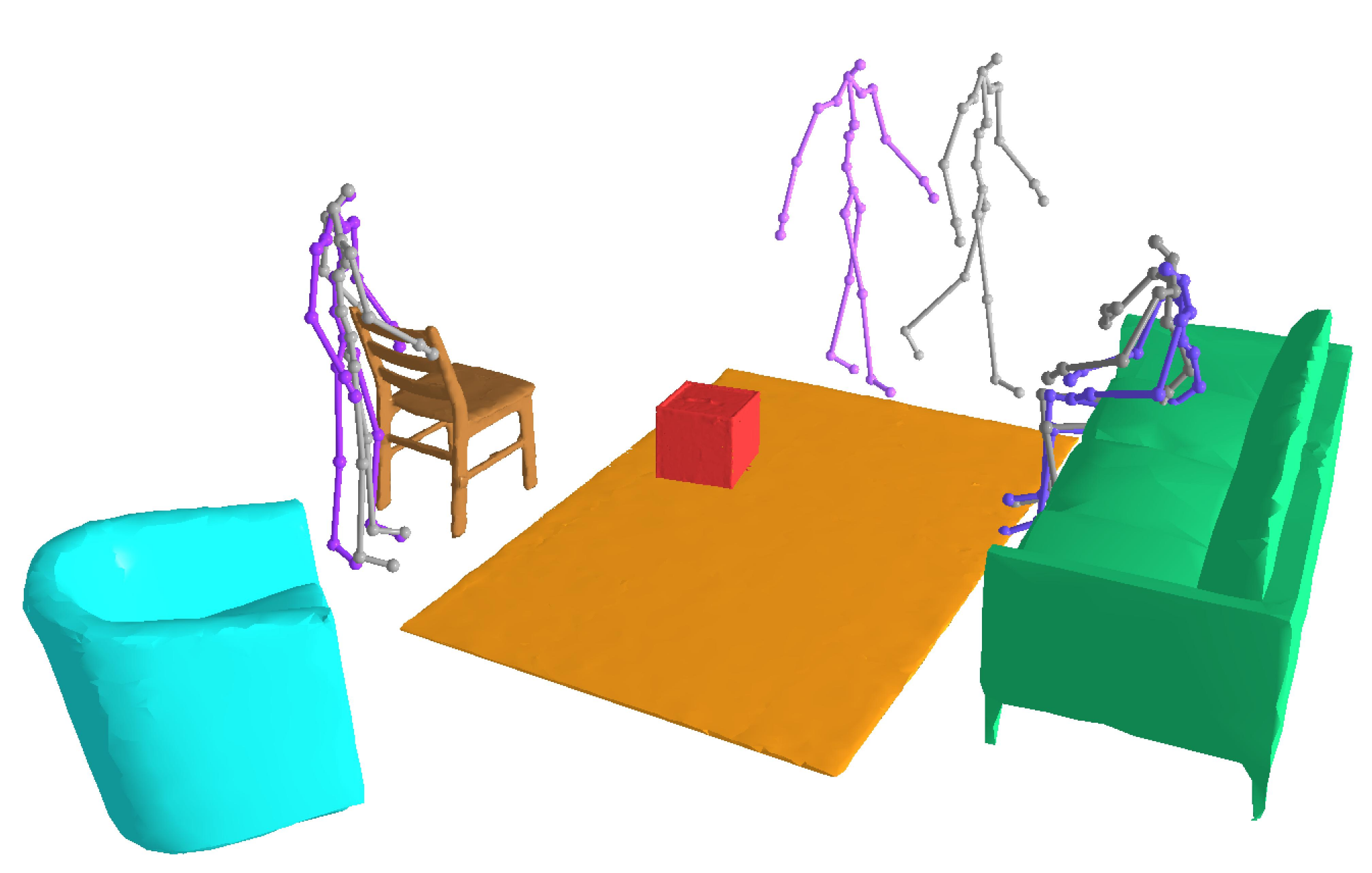} \\
    (a) Frame 29 ($t=1$) & (b) Frame 59 ($t=2s$) & (c) Frame 89 ($t=3s$) \\
  \end{tabular}
    \caption{\textbf{2-second prediction on the HOI-M\textsuperscript{3} dataset.} 
    Purple skeletons represent our predictions, while gray skeletons represent GT.
    The model accurately captures the pose dynamics over the 2-second horizon.}
    \label{fig:sec2hoi1}
\end{figure*}

In column (c) of Fig.~\ref{fig:sec2hoi1}, the central person continues to walk. The predicted motion follows with the correct poses but a little bit slower. In the mean time, the interaction of the left person with the chair continues closely to the real motion. The model correctly identifies the affordance for interaction and also generates a valid walking path.

\subsection{HiK dataset: 10-second prediction}
\begin{figure*}[t]
  \centering
  \setlength\tabcolsep{1pt}
  \begin{tabular}{ccc}
    \includegraphics[width=0.32\linewidth]{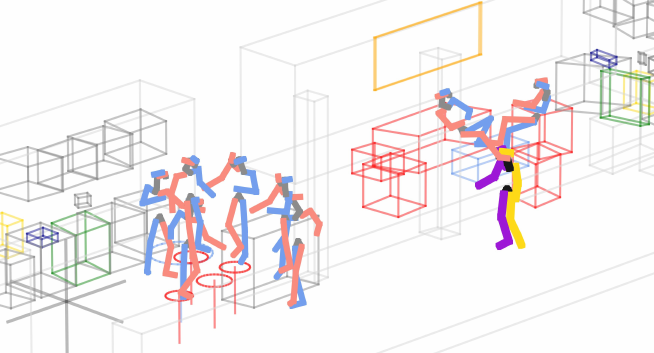} &
    \includegraphics[width=0.32\linewidth]{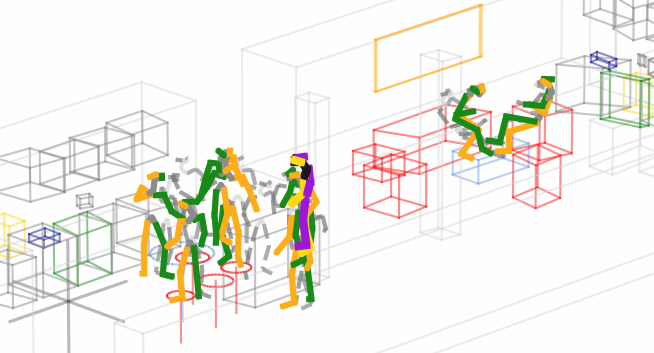} &
    \includegraphics[width=0.32\linewidth]{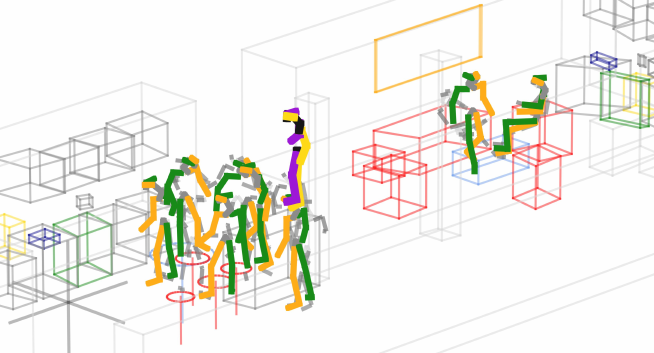} \\
    (a) Last Ground Truth & (b) Prediction Sample 1 & (c) Prediction Sample 2 \\
  \end{tabular}
    \caption{\textbf{Qualitative comparison on the HiK dataset (10-second prediction).} 
    We visualize two stochastic samples for the same scene.
    (a) Last observed GT at $t = 2$ s (frame 49). 
    (b - c) Final predicted frame at $t = 12$ s (frame 299) from two distinct generated samples.
     Visualization scheme follows Fig.~\ref{fig:sec2hik1}. The purple/yellow person walks to different locations in (b) and (c). This demonstrates the model's ability to generate diverse predictions.}
    \label{fig:sec2hik3}
\end{figure*}

In Fig.~\ref{fig:sec2hik3}, we examine the stochastic nature of our model by generating multiple predictions for a single 10-second scenario. While column (a) shows the initial state,  columns (b) and (c) display two different samples generated by our model for frame 299. As observed, the model generates diverse but plausible futures. The purple/yellow person walks to different spatial locations in sample 1 versus sample 2. This confirms that OCSD is not collapsing to a single deterministic mean, but is capable of modeling the natural variance in human motion over long horizons.

\begin{figure*}[t]
  \centering
  \setlength\tabcolsep{1pt}
  \begin{tabular}{ccc}
    \includegraphics[width=0.32\linewidth]{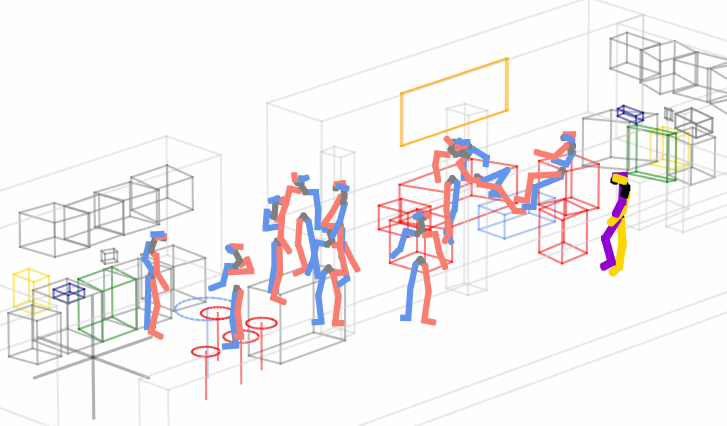} &
    \includegraphics[width=0.32\linewidth]{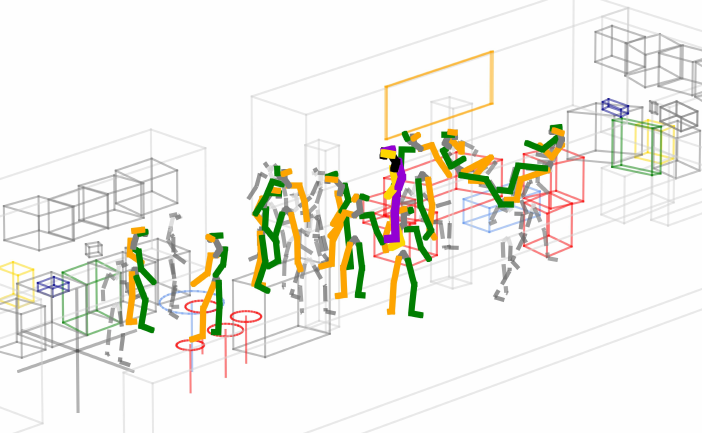} &
    \includegraphics[width=0.32\linewidth]{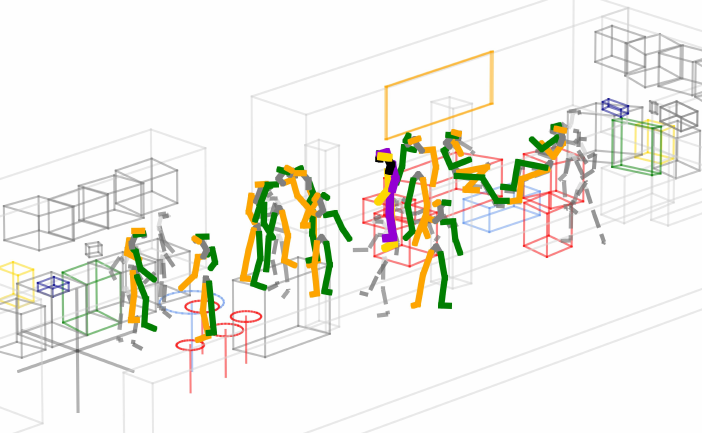} \\
     & (a) OCSD (ours) &  \\
    \includegraphics[width=0.32\linewidth]{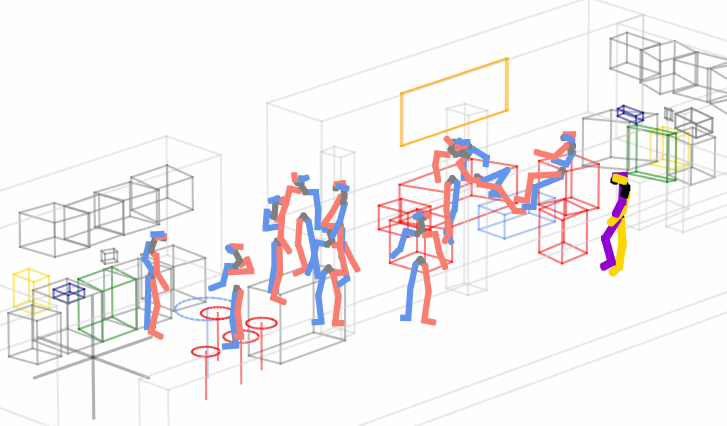} &
    \includegraphics[width=0.32\linewidth]{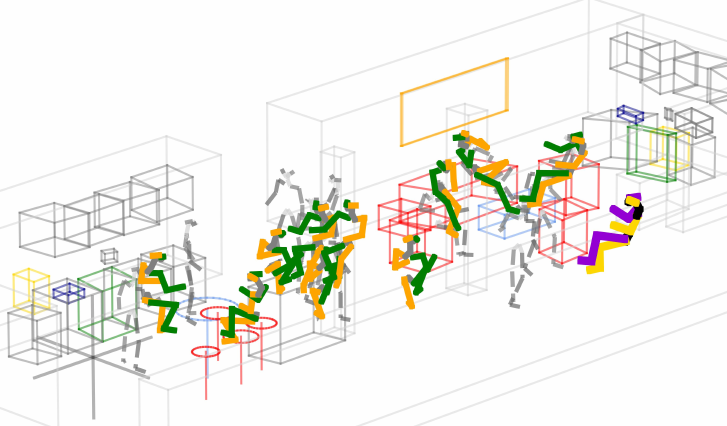} &
    \includegraphics[width=0.32\linewidth]{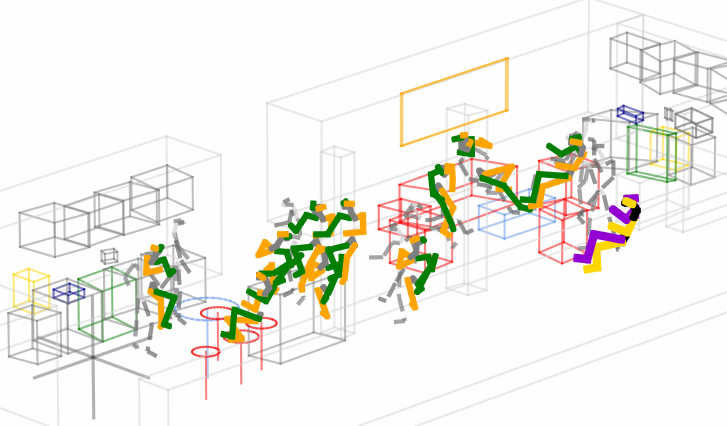} \\
     & (b) SAST &  \\
  \end{tabular}
    \caption{\textbf{Qualitative comparison with SAST (HiK dataset, 10-second prediction).}
     Each row shows the last observed GT at Frame 49 ($t = 2$ sec), prediction at Frame 174 ($t=7$ sec), and prediction at Frame 299 ($t=12$ sec). Visualization scheme follows Fig.~\ref{fig:sec2hik1}. For SAST, the purple/yellow person hallucinates a chair and sits down at a location where there is no chair (row b). Our model produces plausible motion for the persons (row a) over the full $10$-second prediction horizon.
    }
    \label{fig:sec2hik4}
\end{figure*}

\textbf{Comparison with SAST (10-second).} 
We further compare long-term prediction capabilities in Fig.~\ref{fig:sec2hik4}. The SAST model (row b) forecasts that the purple/yellow person sits down at a location where there is no chair. In contrast, OCSD (row a) maintains motion plausibility throughout the $10$-second prediction window.

\section{Limitations}
\label{sec:limits}

\begin{figure*}[t]
  \centering
  \setlength\tabcolsep{1pt}
  \begin{tabular}{ccc}
    \includegraphics[width=0.32\linewidth]{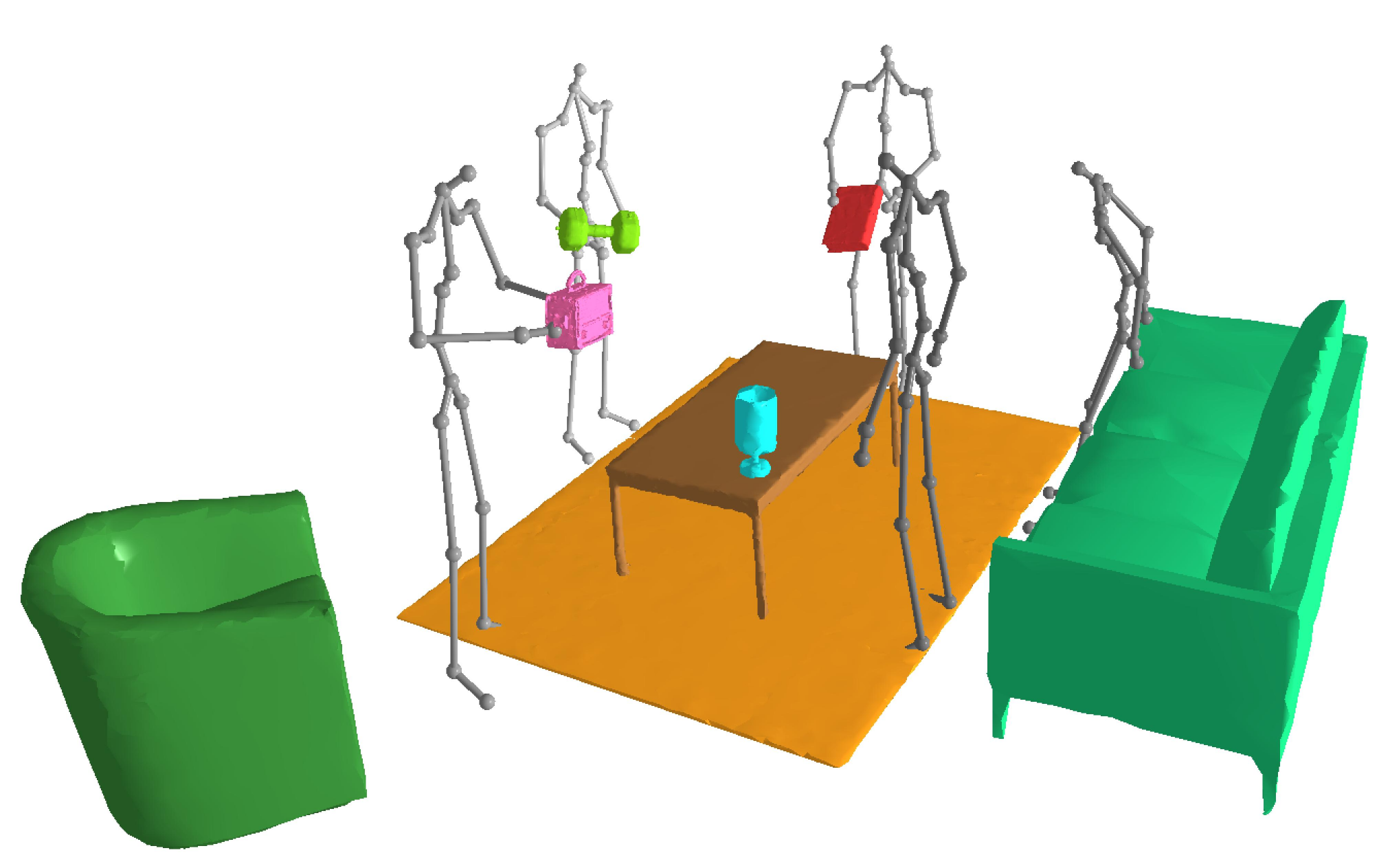} &
    \includegraphics[width=0.32\linewidth]{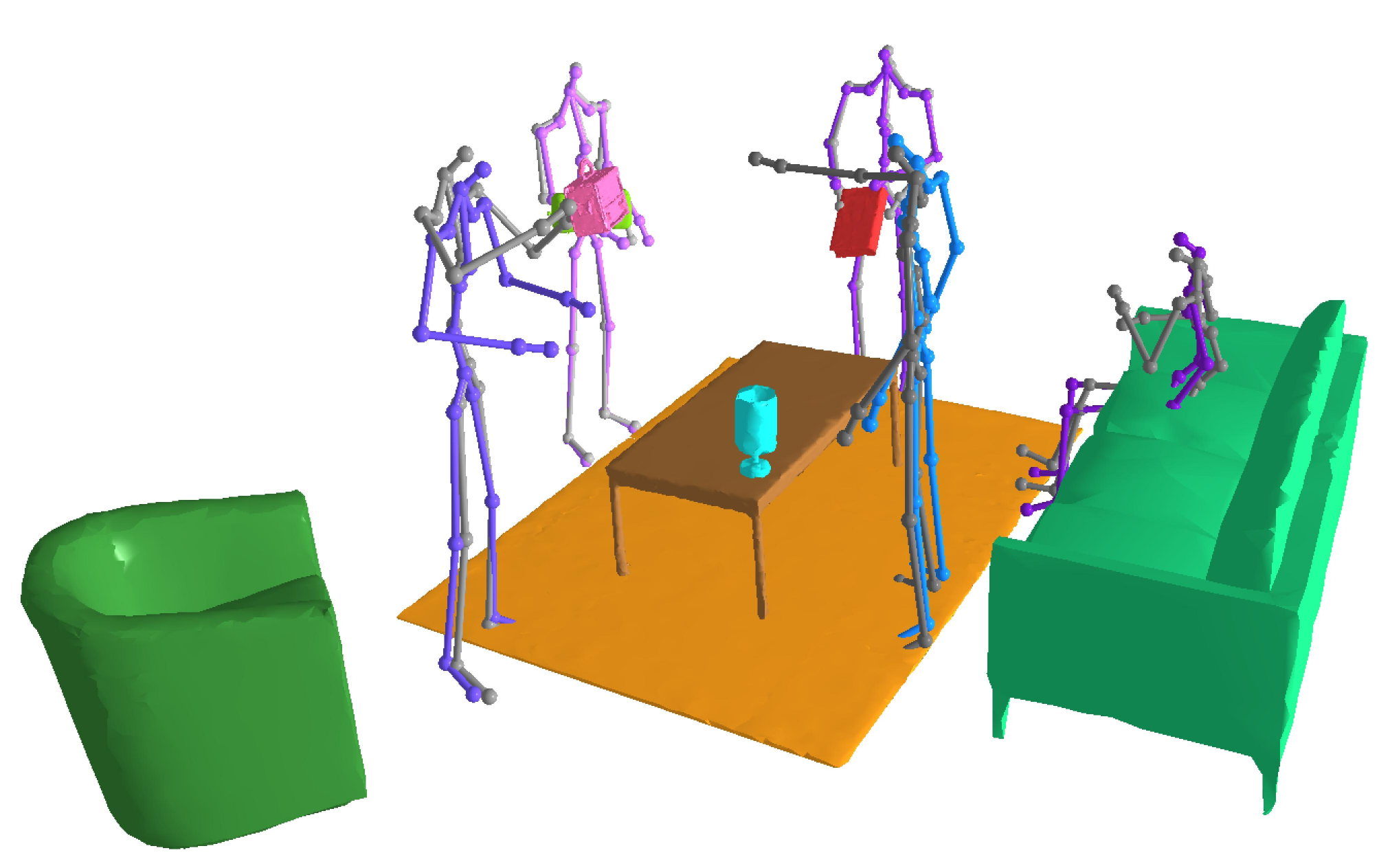} &
    \includegraphics[width=0.32\linewidth]{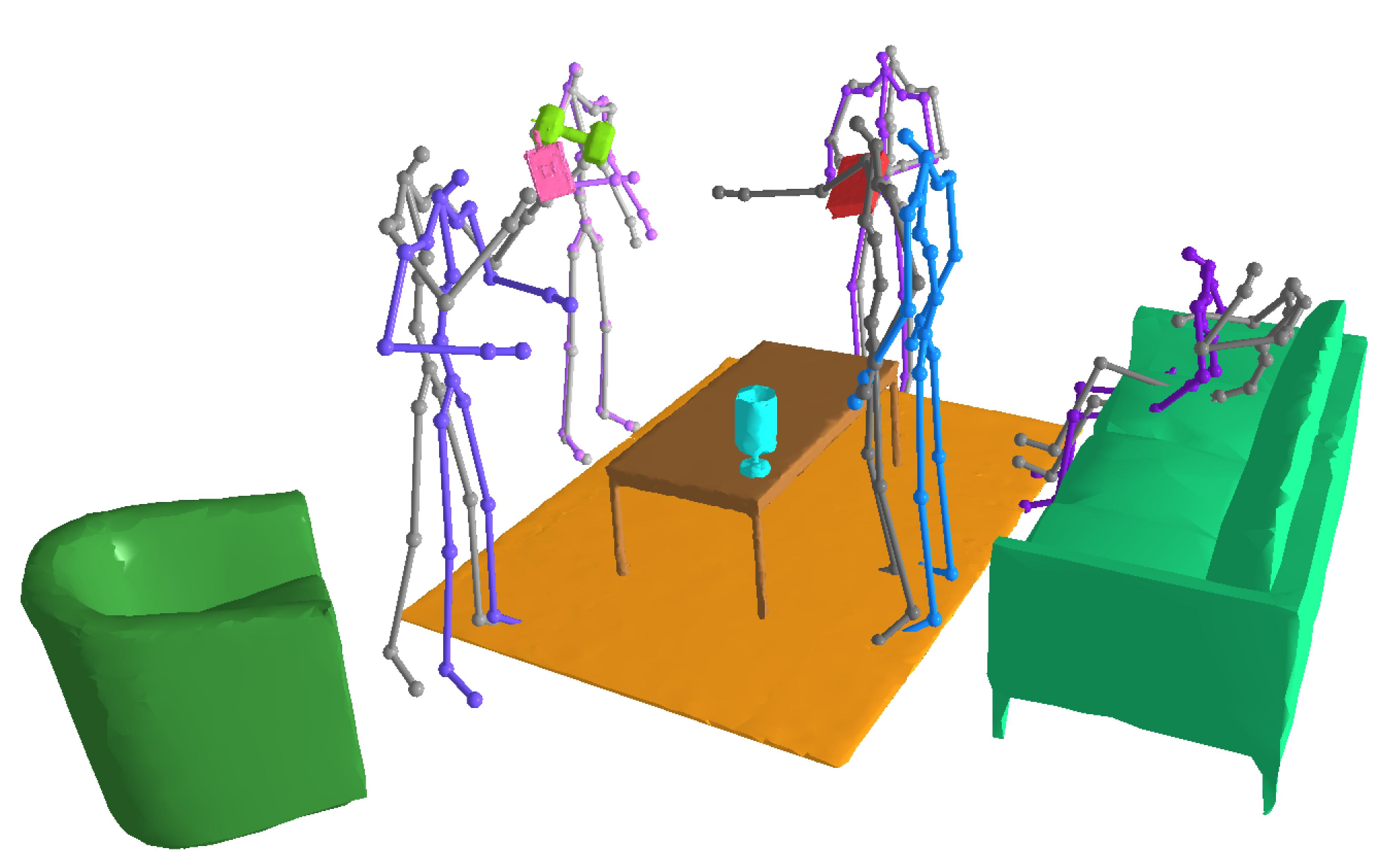} \\
    (a) Frame 29 ($t=1$) & (b) Frame 59 ($t=2s$) & (c) Frame 89 ($t=3s$) \\
  \end{tabular}
    \caption{\textbf{Limitation regarding moving objects.} 
    Purple skeletons represent our predictions, while gray skeletons represent GT.
    Since the future position of the objects are unknown for the model during the prediction interval, predictions deviate from the actual human-object interaction. This is a 2-second prediction on the HOI-M\textsuperscript{3} dataset.}
    \label{fig:limit}
\end{figure*}

The first limitation regards moving objects in the prediction interval. Our model uses the last observed positions of the objects for that interval. This avoids providing the model with oracle knowledge of future object positions, but it means that predictions cannot adapt to objects that move in the future. For example in Fig.~\ref{fig:limit}, the positions of the bag, dumbbell and book change during the prediction interval, and the predicted motions remain anchored to their last observed positions. This causes predictions to deviate from the actual human-object interaction, and is observed in a subset of HOI-M\textsuperscript{3}. HiK is largely unaffected since their objects are mostly static. Addressing this would require jointly predicting the future positions of the objects alongside human motion, a promising direction for future work. 

A second limitation concerns the semantic object representation. Our one-hot encoding for object category information is effective for the closed object vocabularies of HiK and HOI-M\textsuperscript{3}, but it is not suitable for open-vocabulary generalization. Preliminary experiments with CLIP~\cite{radford2021learning} text embeddings of object names did not improve our approach. As future work, vision-language embeddings could be integrated into the model for more generalizable object representations.

\section{Additional Experiment Results}
\label{sec:extabl}
We perform a range of additional ablation studies to further justify the components of our method, report additional metrics focused on motion quality, evaluate the effect of the number of samples on the diversity, and report best-of-K results instead of the average score.
\subsection{Social encoder design}
We validate our social encoder design by showing that our mean pooling outperforms cross-attention for encoding social information in Tab.~\ref{tab:cross}, and that injecting this into the bottleneck gives better results than doing so throughout the network in Tab.~\ref{tab:bottleneck}.
\begin{table}[ht!]
\centering
\caption{\textbf{Social Encoder ablation on HiK.} Replacing mean pooling with cross-attention or integrating social context at multiple U-Net layers instead of the bottleneck increases mean path error (mm) and pose error (mm). }
\label{tab:social}
\begin{subtable}[t]{0.48\linewidth}
\centering
\caption{Mean pooling vs.\ Cross-attention}
\label{tab:cross}
    \begin{tabular}{lcc}
    \toprule
    Method & Path Error$\downarrow$ & Pose Error$\downarrow$ \\
    \midrule
    \rowcolor{blue!20!white}{Mean pooling}   &  \textbf{137.6} & \textbf{88.4}  \\
    Cross-attn  & 142.3 & 91.0 \\
    \bottomrule
    \end{tabular}
\end{subtable}
\hfill
\begin{subtable}[t]{0.48\linewidth}
\centering
\caption{ Bottleneck vs.\ Multi-layer}
\label{tab:bottleneck}
    \begin{tabular}{lcc}
    \toprule
    Social context & Path Error$\downarrow$ & Pose Error$\downarrow$ \\
    \midrule
    \rowcolor{blue!20!white}{Bottleneck}  & \textbf{137.6} & \textbf{88.4} \\
    Multiple layers & 140.2 & 89.9 \\
    \bottomrule
    \end{tabular}
\end{subtable}
\end{table}

\subsection{Context contribution}
\label{appcontext}
While both social and object context have a clear positive effect on the path error, the averaged results in 
%Tab.~\ref{tab:ablation_context_short} 
the main paper
do not show the full picture for the pose error. To demonstrate this, we report the results of 
%Tab.~\ref{tab:ablation_context_short} 
Tab.~4 for three representative activities in Tab.~\ref{tab:activity}.
We find that for the walking activity, where context is important for the path but not necessarily for the pose, including the context slightly increases the pose error. In contrast, for the more static, object-centered social activities, such as interacting with a coffee machine and the surrounding people, the inclusion of this context leads to better predictions. For the path error, context is always helpful.
\begin{table}[H]
\centering
\caption{ \textbf{Context effect per activity on HiK.} Context improves mean path error for all three representative activities. For pose error, benefits provided by context are only evident in the object-centered social activities sink and coffee.} 
\label{tab:activity}
\begin{tabular}{l ccc c ccc}
\hline
Method & \multicolumn{3}{c}{Path Error (mm) $\downarrow$}  & & \multicolumn{3}{c}{Pose Error (mm) $\downarrow$} \\
\cline{2-4} \cline{6-8}
& walking & sink & coffee & & walking & sink & coffee \\
\hline
\rowcolor{blue!20!white}\textbf{OCSD } & \textbf{306.0} & \textbf{57.3} & \textbf{73.6} & &100.5 & \textbf{57.3} & \textbf{58.3}  \\
self + social & 328.3 & 78.3 & 94.0 & & 98.6 & 65.0 & 62.2  \\
self + object& 319.8 & 68.4 & 84.9& & 99.5 & 65.8 & 62.2  \\
only self & 334.1 & 81.4 & 96.0 & & \textbf{98.3} & 63.8 & 62.4 \\
\hline
\end{tabular}
\end{table}

\subsection{U-Net vs. DiT}
We exchange our U-Net backbone for a DiT, keeping the conditioning and training protocol identical. Tab.~\ref{tab:dit} shows that DiTs overfit on the relatively small multi-person motion forecasting datasets with object information and U-Nets remain the better choice. 
\begin{table}[ht!]
\centering
\caption{\textbf{U-Net vs. DiT backbone on HiK.} Diffusion transformer performs worse than U-Net.
}
\label{tab:dit}
\begin{tabular}{l rrrrr r rrrrr}
\hline
Backbone & \multicolumn{5}{c}{Path Error (mm) $\downarrow$} & & \multicolumn{5}{c}{Pose Error (mm) $\downarrow$} \\
\cline{2-6} \cline{8-12}
& 0.5s & 1.0s & 1.5s & 2.0s & mean & & 0.5s & 1.0s & 1.5s & 2.0s & mean \\
\hline
\rowcolor{blue!20!white}\textbf{U-Net}& \textbf{78.9} & \textbf{132.7} & \textbf{197.8} & \textbf{266.9} & \textbf{137.6} & & \textbf{69.8} & \textbf{99.4} & \textbf{112.3} & \textbf{121.9} & \textbf{88.4} \\
\textbf{DiT}  & 88.5 & 148.6 & 223.6 & 294.3 & 154.0 & & 74.7 & 103.2 & 121.3 & 133.7 & 94.5 \\
\hline
\end{tabular}
\end{table}

\subsection{Additional motion quality metrics}
We report four additional metrics on HOI-M\textsuperscript{3} in Table \ref{tab:interpen}: Foot sliding (mm/s), Interpenetration depth (mm), Interpenetration rate, and the Mean Per Joint Velocity Error (MPJVE) (mm/s). We compare our results against ground truth (GT) motion. Foot sliding of predictions is within 5\% of the GT. Interpenetration statistics are likewise close to GT. Note that ground-truth motion has penetrations as well due to approximations of the object surfaces. 
The results show that OCSD does not suffer from strong motion quality artifacts. This is also evident from the NDMS scores, which explicitly compares the velocity-magnitude ratios against real motion, reported in the paper.    
%and are consistent with our highest NDMS, which explicitly scores velocity-magnitude ratios against real motion. 

\begin{table}[ht!]
\centering
\caption{\textbf{Additional motion quality metrics on HOI-M\textsuperscript{3}.} 
OCSD closely matches ground truth foot sliding and penetration statistics.
} 
\label{tab:interpen}
\begin{tabular}{l ccccc}
\hline
    Method \quad & Foot sliding (mm/s) \quad & Interp.depth (mm) \quad & Interp.rate \quad & MPJVE (mm/s) & \\
\hline
GT  & 219.3 & 38.5 & 0.0338 & N/A  \\
OCSD & 208.2 & 42.3 &   0.0343 & 330.2  \\
\hline
\end{tabular}
\end{table}

\subsection{Number of evaluation samples}
Tab.~\ref{tab:umwr} shows that increasing the number of samples used for evaluation from 4 to 16 increases the UMWR, especially for 2-second predictions. 
This confirms that diversity is not saturated at four samples.
\begin{table}[H]
\centering
\caption{\textbf{Effect of sample count on diversity (UMWR) on HiK.} 
Diversity scores increase with more samples. 
}
\label{tab:umwr}
\begin{tabular}{l ccccc}
\hline
Samples & \multicolumn{5}{c}{UMWR  $\uparrow$} \\
\cline{2-6} 
 &2s & 4s & 6s & 8s & 10s \\
\midrule
 4 &  0.47 & 0.45 & 0.45 & 0.44 &0.44  \\
16 & 0.56 & 0.48 & 0.46 & 0.46 & 0.45 \\
\bottomrule
\end{tabular}
%}
\end{table}

\subsection{Best-of-K analysis}
Tab.~\ref{tab:bestofk} reports best-of-K path and pose errors on HiK for K $\in$ $\{1, 5, 10, 20\}$. Results are already good for $K=1$ compared to the baselines, but they get even better for higher values. This supports the previous results, showing that OCSD produces diverse, yet consistently high-quality predictions.  

\begin{table}[H]
\centering
\caption{\textbf{Best-of-K analysis on HiK.} Accuracy improves steadily with K. OCSD produces diverse, yet consistently high-quality predictions.} 
\label{tab:bestofk}
%\resizebox{0.8\textwidth}{!}{
\begin{tabular}{l rrrrr r rrrrr}
\hline
K & \multicolumn{5}{c}{Path Error (mm) $\downarrow$} & & \multicolumn{5}{c}{Pose Error (mm) $\downarrow$} \\
\cline{2-6} \cline{8-12}
& 0.5s & 1.0s & 1.5s & 2.0s & mean & & 0.5s & 1.0s & 1.5s & 2.0s & mean \\
\hline
\textbf{1}& 79.0 & 131.0 & 200.8 & 270.9 & 139.0 & & 69.5 & 101.8 & 115.0 & 121.8 & 89.0 \\
\textbf{5}  & 57.5 & 90.7 & 134.8 & 181.4 & 95.5 & & 54.2 & 75.0 & 87.5 & 93.1 & 68.3 \\
\textbf{10} & 51.8 & 79.2 & 116.7 & 156.1 & 83.5 & & 50.1 & 68.3 & 79.8 & 85.1 & 62.5 \\
\textbf{20}  & 46.5 & 69.3	& 103.0 & 137.4 & 74.0 & & 47.1 & 63.3 & 74.0  & 79.4 & 58.3 \\
\hline
\end{tabular}
%}
\end{table}